\documentclass[]{xiaomiev}

\usepackage[toc,page,header]{appendix}
\usepackage{minitoc}
\usepackage{solarized-light}
\usepackage{booktabs}
\usepackage{multirow}
\usepackage{graphicx}
\usepackage{array}
\usepackage{siunitx,array}
\usepackage[table]{xcolor}
\usepackage{makecell,array,booktabs}
\usepackage{makecell}
\usepackage{framed}

\usepackage{bbding}
\usepackage{hyperref}
\usepackage{amsmath}
\usepackage{amssymb}
\usepackage{placeins}
\usepackage[colorinlistoftodos]{todonotes}
\usepackage{longtable}
\usepackage{hhline}
\usepackage{fancyvrb}
\usepackage{graphicx}
\usepackage[table]{xcolor}
\usepackage{booktabs}
\usepackage{float}
\usepackage{fvextra}
\usepackage{CJKutf8}
\usepackage{multicol}
\usepackage{float}
\usepackage{wrapfig}
\usepackage{placeins}
\usepackage{cleveref}
\usepackage{tablefootnote}
\usepackage{threeparttable}
\usepackage{tabularx}
\usepackage{subcaption}
\usepackage[usestackEOL]{stackengine}
\usepackage{tikz}
\usetikzlibrary{arrows.meta,shapes,shapes.geometric,positioning,fit,calc,decorations.pathreplacing}
\usepackage{pgfplots}
\pgfplotsset{compat=1.18}
\usepgfplotslibrary{groupplots}
\usepackage[numbers]{natbib}
\newcommand{\commentout}[1]{}
\renewcommand{\paragraph}[1]{\noindent\textbf{#1.}\hspace*{1em}}
\usepackage{enumitem}
\usepackage{hyperref}
\setlist[itemize]{leftmargin=15pt}
\usepackage{siunitx,array}
\usepackage{graphicx}
\usepackage{xcolor}
\usepackage{wasysym}

\RequirePackage{xspace}
\makeatletter
\DeclareRobustCommand\onedot{\futurelet\@let@token\@onedot}
\def\@onedot{\ifx\@let@token.\else.\null\fi\xspace}

\def\eg{\textit{e.g}\onedot} 
\def\ie{\textit{i.e}\onedot}

\makeatother

\newcommand{\OneVL}{OneVL}

\title{Xiaomi \OneVL{}: One-Step Latent Reasoning and Planning with Vision-Language Explanation}

\author{Xiaomi Embodied Intelligence Team}
\vspace{-11pt}

\contribution{See the \protect\hyperref[sec:contributions]{\textbf{Contributions and Acknowledgments}} section for a list of contributors.}

\abstract{
Chain-of-Thought (CoT) reasoning has become a powerful driver of trajectory prediction in VLA-based autonomous driving, yet its autoregressive nature imposes a latency cost that is prohibitive for real-time deployment. Latent CoT methods attempt to close this gap by compressing reasoning into continuous hidden states, but consistently fall short of their explicit counterparts. We suggest that this is due to purely linguistic latent representations compressing a symbolic abstraction of the world, rather than the causal dynamics that actually govern driving. Thus, we present \textbf{OneVL} (\textbf{One}-step latent reasoning and planning with \textbf{V}ision-\textbf{L}anguage explanations), a unified VLA and World Model framework that routes reasoning through compact latent tokens supervised by dual auxiliary decoders. Alongside a language decoder that reconstructs text CoT, we introduce a visual world model decoder that predicts future-frame tokens, forcing the latent space to internalize the causal dynamics of road geometry, agent motion, and environmental change. A three-stage training pipeline progressively aligns these latents with trajectory, language, and visual objectives, ensuring stable joint optimization. In inference, the auxiliary decoders are discarded, and all latent tokens are prefilled in a single parallel pass, matching the speed of answer-only prediction. Across four benchmarks, OneVL becomes the first latent CoT method to surpass explicit CoT, delivering superior accuracy at answer-only latency. These results show that with world model supervision, latent CoT produces more generalizable representations than verbose token-by-token reasoning.
\\[2ex]
\textbf{- Project Page:} \href{https://Xiaomi-Embodied-Intelligence.github.io/OneVL}{\textbf{https://Xiaomi-Embodied-Intelligence.github.io/OneVL}}
\\[1ex]
\textbf{- GitHub Repository:} \href{https://github.com/xiaomi-research/OneVL}{\textbf{https://github.com/xiaomi-research/OneVL}}
}

\begin{document}

\maketitle

\begin{figure}[H]
\centering
\vspace{-0.2cm}
\includegraphics[width=0.9999\textwidth]{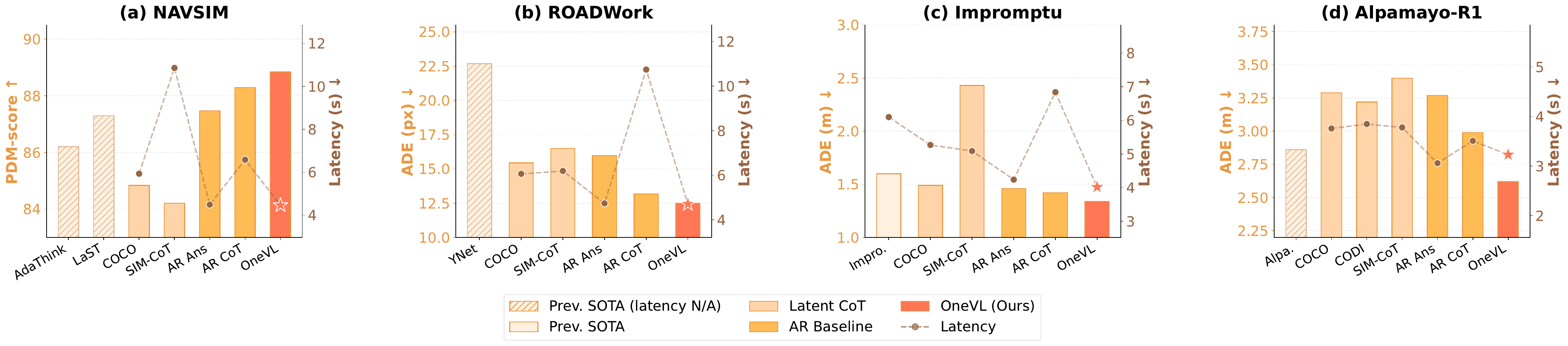}
\caption{\textbf{Accuracy and efficiency comparison across four benchmarks.} Existing latent CoT methods underperform explicit CoT. OneVL is the first to surpass it while matching answer-only prediction latency.}
\label{fig:teaser}
\end{figure}

\clearpage
\onecolumn

\tableofcontents
\newpage

\section{Introduction}
\label{sec:intro}

Vision-Language Models (VLMs)~\cite{gpt4o,bai2023qwenvlversatilevisionlanguagemodel,bai2025qwen2,bai2025qwen3,wang2025internvl3,liu2023visual,liu2024improved,anthropic20253,gemini25pro,guo2025deepseek,seed15,tong2024cambrian,luo2026unleashing,chen2026vilta,chen2025nanovla,wang2025vision,huang2025robotron} have rapidly become a foundational building block for autonomous driving, unifying holistic scene understanding, natural language reasoning, and end-to-end trajectory planning within a single model~\cite{survey_vla4ad,survey_3d_4d_world_models,sima2024drivelm,xu2024drivegpt4,wang2024omnidrive,marcu2024lingoqa,worldlens,yan2025ad-r1, hou2025drivemrp,huang2023fuller}. When further extended to produce action outputs, such as trajectory waypoints or control signals, these models are known as Vision-Language-Action models (VLAs)~\cite{drivingwithllms2024,lastvla,chiimpromptu,recogdrive,hao2025mimoembodiedxembodiedfoundationmodel,fu2025orion,fu2025minddrive,huang2024making,wang2026vggdrive}.

A central driver of recent progress in VLA-based driving is Chain-of-Thought (CoT) reasoning~\cite{adathinkdrive,zeng2025futuresightdrive,lastvla,chiimpromptu}, where the model articulates intermediate reasoning steps before committing to a final trajectory, yielding substantial gains in prediction quality~\cite{survey_vla4ad,wang2025alpamayo}. By explicitly surfacing scene semantics, anticipated agent behaviors, and high-level driving intent, CoT supervision binds predictions into coherent causal chains and markedly reduces planning errors. This success echoes a broader body of LLM CoT research spanning mathematical reasoning~\cite{lightman2023lets}, document understanding~\cite{lu-etal-2025-bounding,lu2023punifiedner,lu2024padellm,fei-etal-2025-advancing,lu2023makes,feng2025dolphin}, code synthesis~\cite{chen2021evaluating}, multimodal QA~\cite{lu2022learn,tang-etal-2025-mtvqa}, RL-based deep reasoning~\cite{guo2025deepseek,jia2026meml}, and test-time scaling~\cite{openai2024o1,snell2024scaling}. A unifying explanation for why CoT works comes from the \emph{compression view of intelligence}~\cite{legg2007universal,deletang2024lm}: under next-token supervision, a model forced to articulate intermediate steps must compress its understanding into structured, generalizable representations rather than memorize shallow input–output mappings.

Yet deploying CoT in real driving systems exposes a sharp tension between interpretability and efficiency. Standard autoregressive~(AR) CoT generation must emit every reasoning token before the trajectory can be produced. This yields inference latency proportional to the chain length, which is far above that of answer-only prediction. In safety-critical real-time settings, this gap is prohibitive. At the same time, explicit CoT chains are strikingly redundant; for example, much of the sequence merely restates context or follows formulaic patterns. This redundancy suggests that the essential reasoning content can be compressed into a far more compact form~\cite{tishby1999information} without sacrificing and even strengthening generalization, since tighter compression forces the model to retain only the causal structure that truly matters for prediction.

\paragraph{Latent CoT and Its Limitations}
A growing line of work pursues exactly this direction, replacing explicit reasoning tokens with compact latent representations \cite{yu2026latent}. COCONUT~\cite{coconut} introduced curriculum learning over latent thought tokens, progressively replacing discrete reasoning steps with continuous vectors. CODI~\cite{codi} extended this with self-distillation, training a student to mimic a teacher's CoT behavior in latent space. SIM-CoT~\cite{simcot} attached a separate text-decoding auxiliary decoder to enable direct text supervision during latent training. However, adapting these methods to VLA-based driving reveals critical shortcomings. COCONUT, CODI, and SIM-CoT were designed for language-only reasoning and make no use of the rich visual structure that defines driving scenes. As a result, their purely linguistic latents prove insufficient for the multimodal reasoning demanded by trajectory prediction, and as Figure~\ref{fig:teaser} shows, every existing latent CoT method underperforms explicit CoT across all benchmarks.

\begin{figure}[!t]
    \centering
    \includegraphics[width=0.95\linewidth]{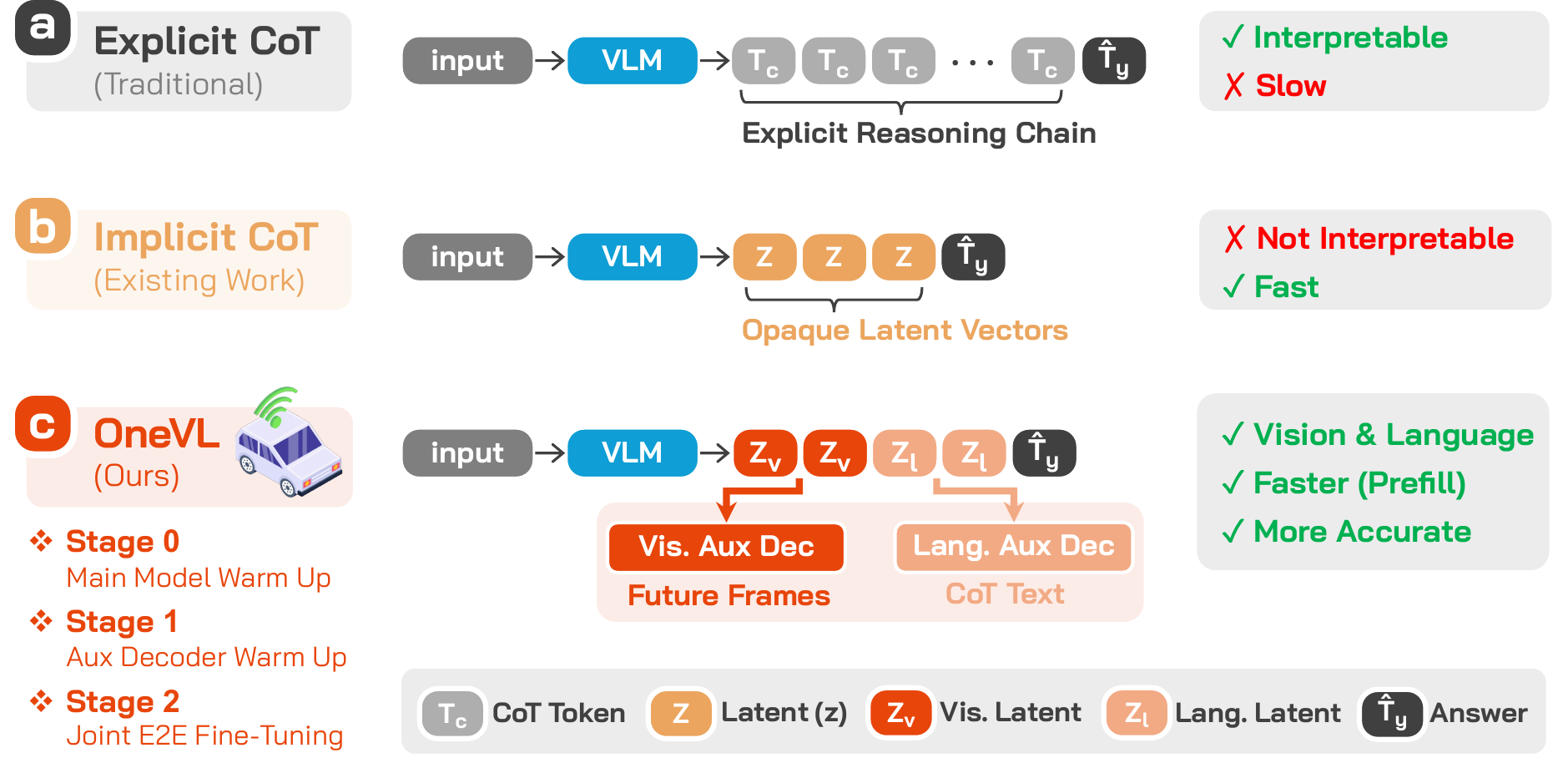}
    \caption{\textbf{Comparison of CoT paradigms.} \textbf{(a)}~\textbf{Explicit CoT:} the model generates a full chain of discrete reasoning tokens before the answer. \textbf{(b)}~\textbf{Implicit CoT:} reasoning is compressed into a small number of opaque latent vectors $\mathcal{Z}$. \textbf{(c)}~\textbf{\OneVL{} (Ours):} two types of latent tokens ($\mathcal{Z}_{v}$, \textcolor{red!70}{red}) and language ($\mathcal{Z}_{l}$, \textcolor{orange!80}{salmon}); during training, dual auxiliary decoders decode these into future-frame visual tokens and CoT text, respectively, providing rich text and world model supervision. During inference, the decoders are discarded, and the latent tokens are prefilled into the prompt context, matching the speed of answer-only prediction while keeping the interpretability of~(a) in both vision and language.}
    \label{fig:cot_comparison}
\end{figure}

More fundamentally, natural language descriptions of driving scenes are inherently abstract. They encode semantic labels rather than the spatiotemporal causal dynamics that actually determine future outcomes. A latent vector that compresses language is therefore compressing a symbolic abstraction of the world, not its underlying causal structure. A further limitation is that in prior methods, latent hidden states are still produced autoregressively (one latent hidden state at a time), leaving inference sequential. We instead aim to generate all latent hidden states in a single step via prefill. Figure~\ref{fig:cot_comparison} contrasts these three paradigms, that is, explicit CoT, prior implicit latent CoT, \OneVL{}, and motivates our design.

\paragraph{OneVL: One-Step Latent Reasoning and Planning with Vision-Language Explanations}

We present \OneVL{}, a framework that overcomes the limitations of prior latent CoT methods through two key innovations. First, we introduce dual-modal auxiliary decoders: a \emph{language auxiliary decoder} that reconstructs human-readable CoT reasoning from compact language latent tokens, and a \emph{visual auxiliary decoder} that predicts anticipated future frames~\cite{emu35,ibq,liureasonplan} from visual latent representations. The visual decoder plays the role of a \emph{world model} auxiliary. By forcing the compressed latents to anticipate what the scene will look like at future time steps, it ensures that the bottleneck encodes genuinely causal scene dynamics, such as agent trajectories, road geometry evolution, and emerging hazards, rather than abstract symbolic summaries. This is precisely the missing ingredient in language-only latent CoT. Future-frame prediction is a concrete compression target that directly reflects the causal structure of the physical world, satisfying the compression view of intelligence in a way that text descriptions alone cannot. The resulting framework simultaneously handles planning, language reasoning, and visual interpretation within a single model.

Beyond interpretability, the dual reconstruction objectives serve a deeper role: they ensure that the compressed latents encode genuinely generalizable structure rather than superficial correlations~\cite{deletang2024lm,tishby2015deep}. If compact latent tokens can be decoded into both coherent language reasoning and plausible future frames, the model has necessarily discovered transferable representations of scene dynamics rather than memorized input-output mappings. Critically, the world model supervision (visual decoder) and the language supervision act as complementary forms of validation. Language grounds the latents in semantic intent, while visual prediction grounds them in physical scene dynamics. Together, they guarantee that the compressed representation satisfies both the semantic and causal requirements of robust trajectory planning.

Second, we design a \emph{Prefill Inference} mechanism. At inference time, the latent tokens (both visual and language) are prefilled into the model's context as fixed prompt inputs, enabling single-pass generation of all latent tokens. This eliminates the iterative latent token generation overhead and achieves inference speed essentially identical to answer-only AR prediction. The resulting model performs one-step latent reasoning (fast inference), vision-language explanation (interpretable reasoning), and finally planning in a unified sequence. Empirically, \OneVL{} not only matches but surpasses explicit AR CoT in trajectory quality, demonstrating that compression, far from being a necessary compromise, is itself a driver of more effective reasoning~\cite{legg2007universal,deletang2024lm}.

\paragraph{Contributions}
The key contributions of this work are summarized as follows:
\begin{itemize}
    \item \textbf{OneVL Framework}: We introduce a latent CoT framework built on the principle that compression drives generalization, and identify a critical gap in prior latent CoT work, that is, purely linguistic latent representations are too abstract to satisfy this principle for planning tasks. We address this with dual-modal auxiliary decoders, a language decoder, and a visual world model decoder that jointly supervise compact latent tokens to encode both linguistic reasoning and future scene dynamics. The world model decoder provides the concrete, causal compression target that language alone cannot supply. A principled three-stage training pipeline progressively aligns the latent bottleneck with trajectory prediction, ensuring that the compressed representations capture causal structure rather than memorized patterns.

    \item \textbf{Superior performance across Four Benchmarks}: \OneVL{} achieves superior performance across many benchmarks. Notably, \OneVL{} is the only latent CoT method that outperforms explicit autoregressive CoT, directly supporting our hypothesis that tighter compression encourages more generalizable reasoning. Ablation studies confirm each component's contribution. Both the visual and language decoders yield consistent performance gains, and the staged training recipe is essential.

    \item \textbf{Prefill Inference}: At inference time, the auxiliary decoders are discarded, and all latent tokens are prefilled into the prompt, enabling single-pass latent CoT reasoning with no iterative overhead. For example, on NAVSIM, the latency matches AR answer-only prediction and is $1.5\times$ faster than explicit autoregressive CoT. On ROADWork, prefill latency is identical to answer-only and $2.3\times$ faster than its explicit counterpart. For real-world deployment, appending an MLP head for producing trajectory further reduces latency to $0.24$s $(4.16$ Hz), just $5.4\%$ of the AR model's latency, offering a practical deployment option.

    \item \textbf{Interpretable Explanations}: The language auxiliary decoder recovers high-quality CoT text from compressed latents, while the visual auxiliary decoder generates spatially coherent future-frame previews, providing both linguistic and visual interpretability.
\end{itemize}

\paragraph{Organization}
The remainder of this paper is organized as follows. \Cref{sec:related} presents related work. \Cref{sec:arch} describes the \OneVL{} architecture in detail, including the main VLM, latent token design, and auxiliary decoders. \Cref{sec:training} elaborates on the three-stage training pipeline and the motivation for each stage. \Cref{sec:evaluation} presents the experimental setup, main results, and ablation studies. \Cref{sec:conclusion} concludes with a discussion of future directions.
\section{Related Work}
\label{sec:related}

\subsection{Implicit and Latent Chain-of-Thought}
\label{sec:related:latent}

Since explicit CoT incurs inference overhead proportional to the length of the reasoning chain, a growing line of work internalizes reasoning into continuous latent representations~\cite{deng2024explicit,cheng2024compressed,wote,su2025token,xu2025softcot,law}. \citet{deng2024explicit} proposed a step-by-step internalization curriculum that progressively replaces each explicit reasoning token with implicit internal computation, training the model to absorb CoT one step at a time. COCONUT~\cite{coconut} generalizes this idea to continuous latent thought tokens through a staged curriculum, enabling breadth-first-like exploration of solution paths entirely within the LLM's hidden state space. Compressed Chain of Thought~\cite{cheng2024compressed} takes a complementary distillation approach, condensing an explicit CoT trace into a small set of dense summary vectors prepended to the input, achieving substantial length reduction at only modest accuracy cost. CODI~\cite{codi} adopts sequence-level self-distillation, training a student model to align its anchor latent hidden state, typically the final hidden representation before the answer, with the teacher model's full chain-of-thought sequence, narrowing the performance gap while preserving efficiency. Token Assorted~\cite{su2025token} offers a flexible middle ground by interleaving discrete text tokens and continuous latent tokens within the same sequence, interpolating between fully explicit and fully implicit reasoning. SIM-CoT~\cite{simcot} identifies a latent instability problem, where representations collapse as the number of latent tokens grows without per-step supervision, and addresses it with a plug-and-play auxiliary decoder that aligns each implicit token with its corresponding explicit reasoning step at training time only.

As discussed, all of these methods were developed for language-only tasks and do not transfer effectively to VLA-based autonomous driving.

\subsection{VLM and VLA for Autonomous Driving}
\label{sec:related:driving}

Beyond the foundational VLM works and CoT-augmented driving models discussed in \Cref{sec:intro}, a parallel line of research has focused on establishing richer evaluation and supervision signals for language-grounded driving~\cite{xie2025drivebench,survey_vla4ad,survey_agentic_world_modeling}. MapLM~\cite{cao2024maplm} introduced a large-scale benchmark specifically targeting map and traffic scene understanding, probing whether VLMs can parse structured road topology from sensor data. \citet{ding2024holistic} augmented VLMs with bird's-eye-view feature injection, enabling holistic scene understanding that fuses camera and top-down spatial context within a single multimodal model. Complementary efforts have explored corner-case evaluation~\cite{chen2025automated} and risk localization~\cite{malla2023drama} to stress-test VLM reasoning under rare or safety-critical scenarios.

Closer to trajectory prediction, recent VLA models pair language reasoning with waypoint or action outputs \cite{liu2026driveworld-vla,liu2025guideflow,zhang2025nava}. DriveVLA-W0~\cite{drivevlaw0} employs world modeling to generate dense self-supervised signals that amplify data scaling laws in VLA-based driving. AdaThinkDrive~\cite{adathinkdrive} introduces adaptive CoT for driving decisions, LaST-VLA~\cite{lastvla} trains a large vision-language-action model on driving data, and Alpamayo-R1~\cite{wang2025alpamayo} explicitly bridges reasoning traces with long-tail action prediction. \OneVL{} builds on these foundations by addressing the latency cost of explicit CoT through dual-modal latent supervision and prefill inference, delivering competitive performance without sacrificing interpretability.

\subsection{World Modeling for Autonomous Driving}
\label{sec:related:world model}

The concept of the \emph{world model} originates from model-based reinforcement learning, where it seeks to emulate human cognitive processes and predict the effects of actions on environmental evolution~\cite{dreamv1,worldmodels,robodrive_challenge_2024}, particularly in 3D and 4D spaces~\cite{survey_3d_4d_world_models,bian2025dynamiccity,liang2026lidarcrafter,xu2025u4d,kong2025lasermix++,kong2026largead}. To further enhance spatial reasoning, several approaches incorporate advanced perception frameworks \cite{wang2025vggt,zuo2025dvgt,zuo2026dvgt,xie2025benchmarking,xu2024gaussianpretrain} to achieve a more robust understanding of 3D environments. With advances in video generation and the introduction of the Joint Embedding Predictive Architecture by \citet{assran2023self}, the scope and applications of world models have broadened considerably~\cite{gaia-1,terver2025drives,dong2026lcvn,hu2026navthinker,dong2026uniwm}. In autonomous driving, world models are typically applied to three ends: data generation, closed-loop evaluation, and representation learning~\cite{wm_survey_ad,survey2,survey_3d_4d_world_models,worldlens,tian2025simscale}.

For data generation, Cosmos~\cite{azzolini2025cosmos} integrates multimodal inputs such as text, images, videos, and motion signals to synthesize consistent data for training robotic and autonomous driving systems. For closed-loop evaluation, DICC~\cite{DICC} leverages generative world models to produce realistic driving images and performs adversarial evaluation on end-to-end driving systems to improve safety and robustness. Similarly, AD-R1~\cite{yan2025ad-r1} takes advantage of the high physical fidelity of world models, employing them as interactive simulators for reinforcement learning and thereby reducing safety violations in challenging scenarios. For representation learning, DriveVLA-W0~\cite{drivevlaw0} incorporates future temporal information from a world model to improve trajectory planning, and DynVLA~\cite{DynVLA} reduces redundancy in generated images by modeling inter-frame similarity, achieving lower inference latency while maintaining competitive performance.

In contrast, \OneVL{} uses short-horizon future visual token prediction as a training-only world model auxiliary paired with compressed latent CoT inside a single VLA. This auxiliary guides the bottleneck toward causal scene dynamics precisely where language-only implicit CoT is insufficient (\Cref{sec:intro}), and is then discarded at inference so that prefilled latents yield answer-only latency. Prior work emphasizes data generation, simulators, or separate representation stacks, rather than this joint certification of language and visual latent bottlenecks.
\section{Model Architecture}
\label{sec:arch}

\begin{figure}[!t]
    \centering
    \includegraphics[width=\linewidth]{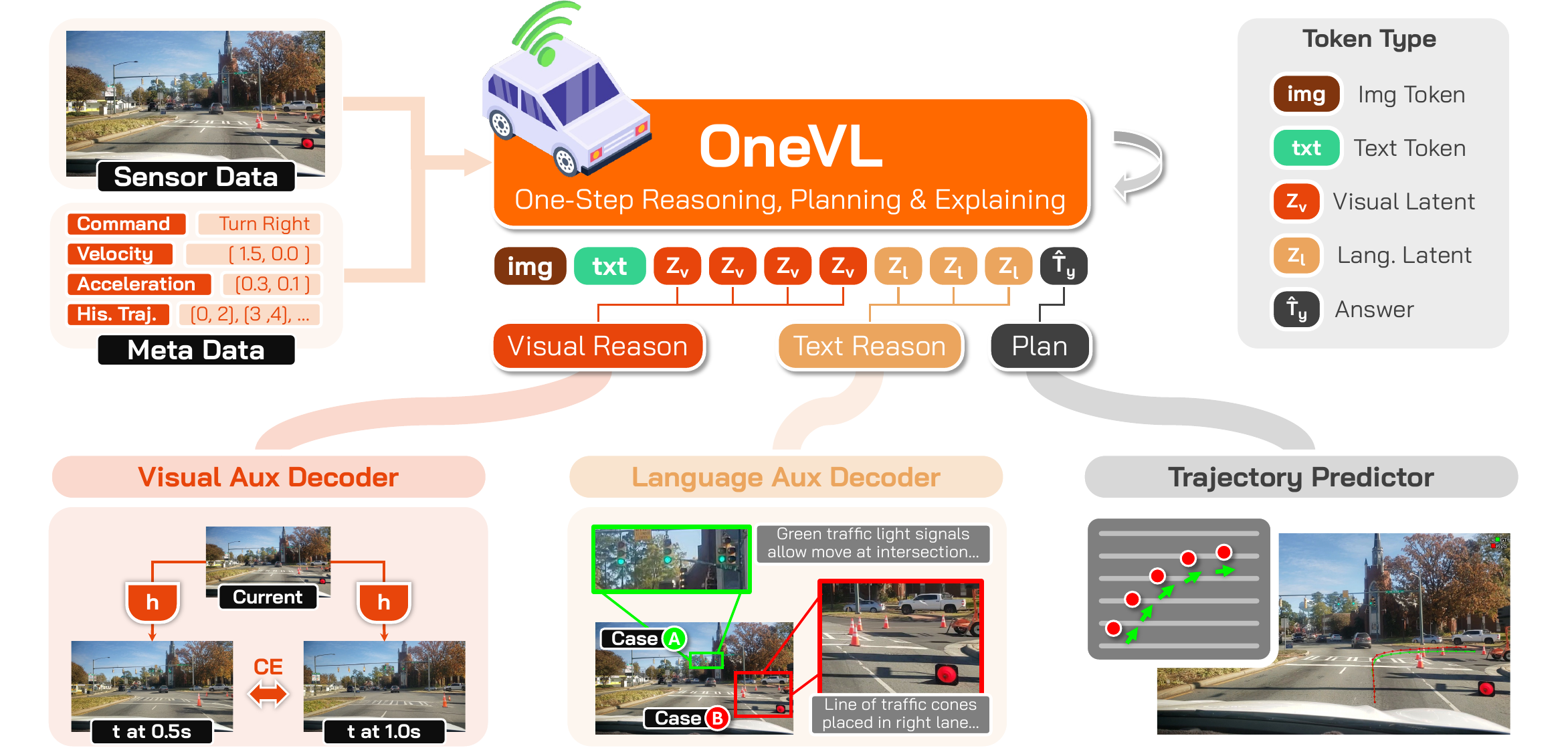}
    \caption{\textbf{\OneVL{} architecture.} An image and structured text prompt (ego state, command, historical trajectory) are fed into the VLM. The output hidden states shown below the VLM, contains image tokens ($\mathcal{T}_{v}$), text tokens ($\mathcal{T}_{l}$), visual latent tokens ($\mathcal{Z}_{v}$), language latent tokens ($\mathcal{Z}_{l}$), and trajectory answer tokens ($\hat{\mathcal{T}}_{y}$). During training, hidden states $\mathcal{H}_{v}$ and $\mathcal{H}_{l}$ at the latent positions are routed to two auxiliary decoders: the Visual Aux.\ Decoder (left) directly predicts future-frame visual tokens at $0.5\mathrm{s}$ and $1.0\mathrm{s}$ ($\mathcal{L}_{v}$), and the Language Aux.\ Decoder (right) predicts chain-of-thought reasoning ($\mathcal{L}_{l}$). During inference, both decoders are discarded; latent tokens are prefilled into the prompt, matching the answer-only AR prediction latency.}
    \label{fig:architecture}
\end{figure}

\OneVL{} augments a pretrained VLM with a compact latent token interface and dual auxiliary decoders for multimodal explanation. Figure~\ref{fig:architecture} gives a complete overview. We describe each component in detail below.

\subsection{Main Vision-Language Model}
\label{sec:arch:main}

The backbone of \OneVL{} is Qwen3-VL-4B-Instruct~\cite{bai2025qwen3}, a VLM that processes interleaved image and text inputs. The model consists of three standard components: Vision Encoder (ViT), Visual Projector (MLP Aligner), and Large Language Model (LLM). All three components are initialized from the Qwen3-VL-4B-Instruct checkpoint and remain fully trainable in Stages 0~(\Cref{sec:training:stage0}) and 2~(\Cref{sec:training:stage2}). The backbone is primarily optimized via a standard next-token prediction objective, applying a cross-entropy loss 
($\mathcal{L}_{c}$) 
to both the trajectory answers and the latent reasoning tokens introduced below.

\subsection{Latent Token Design}
\label{sec:arch:latent}

A critical design decision in \OneVL{} is the introduction of specialized latent tokens that serve as compact carriers of implicit reasoning. We define two classes of latent tokens:

\textbf{Language Latent Tokens} ($\langle|$\texttt{latent}$|\rangle$): A fixed-length sequence of $\mathcal{C}_{t} = 2$ language latent tokens, framed by start and end delimiters ($\langle|$\texttt{start-latent}$|\rangle$ and $\langle|$\texttt{end-latent}$|\rangle$). These tokens are placed in the assistant response before the trajectory answer, occupying the position where explicit CoT reasoning would appear in standard AR models. The hidden states extracted at these token positions after LLM processing encode the model's implicit language-grounded reasoning.

\textbf{Visual Latent Tokens} ($\langle|$\texttt{latent-vis}$|\rangle$): A fixed-length sequence of $\mathcal{C}_{v} = 4$ visual latent tokens, similarly delimited (\ie, $\langle|$\texttt{start-latent-vis}$|\rangle$ and $\langle|$\texttt{end-latent-vis}$|\rangle$), placed before the language latent tokens in the response. These tokens are designed to encode spatial and temporal visual reasoning about the future scene state.

Both sets of latent tokens serve as reasoning carriers whose hidden states at the VLM output layer are fed into auxiliary decoders. Note that during implementation, we found that adding dedicated special tokens (\eg, \texttt{<|latent-vis|>}) to the VLM vocabulary causes performance degradation. Instead, we represent latent tokens using the original vocabulary. Concretely, the $\mathcal{C}_{v}$ visual latent tokens are realized as 35 tokens, and the $\mathcal{C}_{t}$ language latent tokens are realized as 20 tokens.

\subsection{Language Auxiliary Decoder}
\label{sec:arch:lang}

The language auxiliary decoder $\mathcal{D}_l$ aims to recover human-readable CoT reasoning text from the compact language latent hidden states.

\textbf{Input Construction.} For each training sample, the language latent tokens in the main model produce hidden states $\mathcal{H}_{l} \in \mathbb{R}^{\mathcal{C}_t \times d}$, where $d$ is the LLM hidden dimension. We additionally supply the current-frame ViT patch embeddings $\mathcal{V} \in \mathbb{R}^{N_v \times d}$ from the backbone, where $N_{v}$ denotes the length of vision tokens after ViT embedding. An MLP layer maps both branches into the auxiliary decoder's embedding space; we then form the multimodal input by concatenation:
\begin{equation}
  \mathcal{Z}_{l} = \Big[\text{W}_l(\mathcal{V}),\; \text{W}_l(\mathcal{H}_{l})\Big],
\end{equation}
where $\text{W}_l$ is MLP (with dimensions chosen so that $\mathcal{Z}_{l}$ matches the LLM input dimension). The tensor $\mathcal{Z}_{l}$ is fed into $\mathcal{D}_l$.

\textbf{Training Objective.} The language auxiliary decoder is trained to predict the ground-truth CoT reasoning text $\mathcal{T}_{{y}_{_t}}$ given $\mathcal{Z}_{l}$, that is:
\begin{equation}
  \mathcal{L}_l = -\sum_{i=1}^{\vert\mathcal{T}_{{y}_{_t}}\vert} \log P_{\mathcal{D}_l}\left(\mathcal{T}_{{y}_{_t},i} \mid \mathcal{Z}_{l}, \mathcal{T}_{{y}_{_t},<i}\right)~.
\end{equation}
This cross-entropy loss encourages the main model's language latent tokens to encode semantically rich information about the driving scene that is decodable as natural language reasoning.

\subsection{Visual Auxiliary Decoder}
\label{sec:arch:vis}

The visual auxiliary decoder $\mathcal{D}_v$ aims to predict anticipated future-frame visual tokens.

\textbf{Motivation.} Autonomous driving is inherently a spatial-temporal prediction task. Future frame visual tokens, which represent what the driving scene will look like at near-term horizons, are a natural target for learning the visual latent representations. This visual prediction objective serves as a world model auxiliary, supplementing language-only latent CoT. This task acts as a rigorous test of generalization, as predicting unseen configurations requires a robust causal model rather than pattern memorization. By combining visual and language decoders, the framework supervises latents in both physical dynamics and semantic intent, imposing a multi-modal constraint that captures the shared causal structure of the environment.

\textbf{Input Construction.} Let $\mathcal{V} \in \mathbb{R}^{N_v \times d}$ denote the ViT embeddings from the current frame (extracted from the main model's visual encoder), and let $\mathcal{H}_{v} \in \mathbb{R}^{\mathcal{C}_{v} \times d}$ denote the visual latent token hidden states from the main model. Let $\text{W}_v \in \mathbb{R}^{d \times d}$ be the MLP layer. The visual auxiliary decoder receives the concatenation:
\begin{equation}
  \mathcal{Z}_v =\Big[\text{W}_v(\mathcal{V}),\; \text{W}_v(\mathcal{H}_{v})\Big]~.
\end{equation}

This conditioning on both the current visual context and the latent state allows the decoder to perform conditioned future-frame prediction.

\textbf{Visual Tokenizer and Vocabulary Extension.}
To represent images as discrete token sequences, we adopt the IBQ (Index Backpropagation Quantization) visual tokenizer~\cite{ibq}. We use the Emu3.5 tokenizer~\cite{emu35,ibq} with a codebook of 131,072 discrete visual codes. The images are resized to a maximum resolution of 512x512 pixels.
To integrate this visual vocabulary into \OneVL{}, the Qwen3-VL-4B base vocabulary is extended by 131,072 additional visual token IDs. The visual token sequences for training are constructed offline by running the IBQ tokenizer over the ground-truth future frames from the dataset, requiring no additional forward passes during training.

\textbf{Training Objective.} Let $\mathcal{T}_{y_{v}} =\left[\mathcal{T}_{y_{v}, 1},\, \mathcal{T}_{y_{v},2}\right]$ be the concatenated discrete visual token sequence for the future frames at time steps $\mathcal{T}_{y_{v}, 1}$ ($+0.5\text{s}$) and $\mathcal{T}_{y_{v},2}$ ($+1.0\text{s}$). The visual loss is:
\begin{equation}
  \mathcal{L}_v = -\sum_{t=1}^{\vert\mathcal{T}_{y_{v}}\vert} \log P_{\mathcal{D}_v}\!\left(\mathcal{T}_{y_{v},t} \mid \mathcal{Z}_{v}, \mathcal{T}_{y_{v},<t}\right)~.
\end{equation}

\subsection{Combined Training Objective}
\label{sec:arch:loss}

The total training loss $\mathcal{L}$ is a weighted sum of three components:
\begin{equation}
  \label{eq:total_loss}
  \mathcal{L} = \mathcal{L}_{c} + \lambda_{l}  \mathcal{L}_{l} + \lambda_{v} \mathcal{L}_{v}~,
\end{equation}
where $\mathcal{L}_{c}$ is the main model's cross-entropy loss, $\lambda_l = 1.0$ is the language explanation loss weight, and $\lambda_v = 0.1$ is the visual explanation loss weight. The lower weight on $\mathcal{L}_v$ reflects that visual token reconstruction is a harder task, and a smaller weight prevents it from dominating the training signal.

\subsection{Prefill Inference}
\label{sec:arch:inference}

At inference time, the auxiliary decoders are discarded. The key efficiency insight is that the latent tokens, both visual and language, can be prefilled into the prompt context as fixed token sequences, because their specific vocabulary identities have been seen by the model during training.

Concretely, the inference prompt is constructed as:
\begin{equation}
  \resizebox{\dimexpr\linewidth-4em}{!}{$\displaystyle
  \footnotesize
  [\text{System},\, \text{User query},\,
   \langle|\texttt{start-latent-vis}|\rangle,\,
   \underbrace{\langle|\texttt{latent-vis}|\rangle \cdots}_{\mathcal{C}_{v}},\,
   \langle|\texttt{end-latent-vis}|\rangle,\,
   \langle|\texttt{start-latent}|\rangle,\,
   \underbrace{\langle|\texttt{latent}|\rangle \cdots}_{\mathcal{C}_{t}},\,
   \langle|\texttt{end-latent}|\rangle]
  $}~.
\end{equation}

All latent tokens are included in the \emph{prefill} phase rather than the \emph{decode} phase. Since modern transformers~\cite{vaswani2017attention} process the entire prefill in parallel, these additional tokens add negligible overhead compared to sequential autoregressive generation. The model then generates only the trajectory tokens autoregressively. This yields inference latency nearly identical to answer-only AR prediction, while the main model's processing of the prefilled latent tokens still implicitly activates the reasoning pathways learned during training. To conclude, the model outputs:
\begin{itemize}
    \item Trajectory prediction: The primary output—future waypoints for autonomous driving.
    \item Language explanation (optional, via aux decoder): Human-readable CoT reasoning describing the model's interpretation of the scene and its driving decision rationale.
    \item Visual explanation (optional, via visual aux decoder): future frame visual tokens, providing a spatial preview of the predicted scene evolution.
\end{itemize}

Items 2 and 3 are available during post-hoc explanation generation (\eg, for human-in-the-loop~\cite{lu-etal-2022-rationale} debugging, safety auditing, or human-robot interaction), while item 1 is always generated during inference.
\section{Three-Stage Training Pipeline}
\label{sec:training}

Training \OneVL{} presents a unique optimization challenge. The main VLM, the language auxiliary decoder, and the visual auxiliary decoder must all be jointly optimized, yet they have fundamentally different learning objectives and start from different relative states of alignment. 

We address this challenge through a training pipeline consisting of a preliminary self-supervised pretraining step followed by three main stages, each with a clear purpose. The configuration is summarized in Table~\ref{tab:training_config}. 

\subsection{Preliminary: Visual Auxiliary Decoder Self-Supervised Pretraining}
\label{sec:training:pre}

\paragraph{Motivation}
Before integrating the visual auxiliary decoder into the full \OneVL{} pipeline, we first pretrain it independently as a future-frame generator. The intuition is straightforward. Asking the decoder to immediately predict future frames conditioned on latent tokens that carry no information yet (early in training) is an ill-posed task that impedes learning. Instead, we first train the decoder with a strong unconditional prior—given the current-frame ViT embeddings, predict what the scene will look like at the next two timestamps—before introducing the latent conditioning signal. This preliminary stage is conceptually analogous to self-supervised video prediction, where the decoder learns purely from visual observations without any reasoning supervision. 

\paragraph{Training Objective}
The visual auxiliary decoder $\mathcal{D}_v$ receives only the current-frame ViT embeddings $\mathcal{V}$ (projected via $\text{Proj}_v$) as input—the visual latent token hidden states $\mathcal{H}_{v}$ are absent at this stage (the main model is not yet connected). Using the same concatenated target $\mathcal{T}_{y_{v}} =\left[\mathcal{T}_{y_{v}, 1},\, \mathcal{T}_{y_{v},2}\right]$ defined in~\Cref{sec:arch:vis}, the pretraining loss is:
\begin{equation}
  \mathcal{L}_{p} = -\sum_{t=1}^{\vert\mathcal{T}_{y_{v}}\vert} \log P_{\mathcal{D}_v}\!\left(\mathcal{T}_{y_{v}, t} \mid \mathcal{V},\, \mathcal{T}_{y_{v}, <t}\right)~.
\end{equation}

\paragraph{From Unconditioned to Action-Conditioned Generation World Model}
After pretraining, the decoder has learned a robust prior for visual dynamics, enabling it to predict plausible future frames from the current scene alone. This component functions as the model's implicit \textit{world model}, capturing the underlying rules of visual evolution. When it is subsequently connected to the main model, the visual latent tokens $\mathcal{H}^v$ are introduced as an additional conditioning signal alongside the ViT embeddings. Since these latent tokens encode the driving agent's planned action—derived from the main model's reasoning—the decoder effectively transitions from unconditioned next-frame generation to action-conditioned rollouts of the world model. This framing provides a principled interpretation where the visual latent tokens serve as a compact, actionable representation that steers the world model's predictions.

\subsection{Stage 0: Main Model Warmup}
\label{sec:training:stage0}

\paragraph{Motivation}
The fundamental prerequisite for auxiliary decoders to provide meaningful supervision is that the main model's latent tokens (\ie, \texttt{<|latent-vis|>} or \texttt{<|latent|>}) carry information that is semantically aligned with reasoning content. Without a targeted warmup, these tokens would not produce meaningful hidden states that auxiliary decoders can decode.

Stage~0 addresses this by training the main VLM end-to-end on the trajectory prediction task, with latent tokens embedded in each training sample's assistant response. The model learns to:
\begin{itemize}
    \item Predict accurate trajectories: The main CE loss $\mathcal{L}_{\text{CE}}$ over the latent tokens and trajectory answer tokens ensures the model develops a strong base prediction capability.
    \item Develop meaningful latent representations: By contextualizing the latent tokens within the prompt-response structure alongside trajectory targets, the model naturally learns to use the latent positions to encode intermediate representations useful for trajectory prediction. Besides, the attention mechanism allows trajectory tokens to attend to latent token positions, establishing the information routing pathways that the auxiliary decoders will later exploit.
\end{itemize}

\subsection{Stage 1: Auxiliary Decoder Warmup}
\label{sec:training:stage1}

\paragraph{Motivation}
With the main model producing stable, meaningful latent representations (as established in Stage~0), Stage~1 focuses exclusively on training the auxiliary decoders to align with these representations. Crucially, we freeze the main model during this stage, ensuring that the auxiliary decoders optimize against a consistent semantic distribution. By maintaining this stability, the decoders can more effectively internalize the mapping from fixed latent features to visual and language reasoning. To conclude, Stage~1 trains:
\begin{itemize}
    \item Language auxiliary decoder $\mathcal{D}_l$: Trained to decode the language CoT reasoning text and fine-tuned with $\mathcal{L}_l$ against the ground-truth reasoning annotations.
    \item Visual auxiliary decoder $\mathcal{D}_v$: Trained to predict two future frames with $\mathcal{L}_v$.
\end{itemize}

\subsection{Stage 2: Joint End-to-End Fine-tuning}
\label{sec:training:stage2}

\paragraph{Motivation}
Finally, Stage~2 jointly fine-tunes all three model components with the combined loss $\mathcal{L}$ (Eq.~\ref{eq:total_loss}).
The gradients from $\mathcal{L}_l$ and $\mathcal{L}_v$ now flow back into the main model, directly shaping the latent representations to simultaneously serve trajectory prediction, language explanation, and visual prediction objectives. This creates a virtuous cycle:
\begin{itemize}
    \item The richer latent representations enable the main model to make better trajectory predictions (as the latent tokens carry more useful intermediate representations).
    \item The auxiliary decoders adapt to the updated latent representations, improving their explanation quality.
\end{itemize}

This joint optimization is possible in Stage~2 precisely because both the main model and the auxiliary decoders are already well-initialized. Ablation studies (see~\Cref{sec:eval:ablation}) confirm that skipping three-stage training leads to degraded performance.
\section{Experiments}
\label{sec:evaluation}

In this section, we present a comprehensive quantitative evaluation of \OneVL{} across four benchmarks, followed by ablation studies and analyses that isolate the source of each performance gain.

\subsection{Datasets}
\label{sec:dataset}

We evaluate \OneVL{} on four complementary benchmarks: \textbf{NAVSIM}~\cite{dauner2024navsim}, \textbf{ROADWork}~\cite{roadwork}, \textbf{Impromptu}~\cite{chiimpromptu}, and \textbf{Alpamayo-R1}. These datasets are chosen because they have been shown to be effective in settings where CoT reasoning is employed, or because they provide sufficient labels to construct explicit reasoning traces~\cite{adathinkdrive,lastvla,chiimpromptu}.

\begin{itemize}
    \item \textbf{NAVSIM}~\cite{dauner2024navsim} is a large-scale autonomous driving benchmark derived from nuPlan~\cite{karnchanachari2024towards} driving logs, providing real-world data for non-reactive simulation-based planning evaluation.

    \item \textbf{ROADWork}~\cite{roadwork} targets autonomous navigation in road construction zones, featuring temporary signage, non-standard lane configurations, dynamic obstructions such as cones and barriers, and worker presence. These are all scenarios that remain underrepresented in standard driving benchmarks.

    \item \textbf{Impromptu}~\cite{chiimpromptu} is a large-scale vision-language-action benchmark distilled from eight open driving datasets~\cite{IDD,KITTI,Argoverse,waymo,caesar2020nuscenes,dauner2024navsim,noce,Mapillary}. It focuses on four types of unstructured corner-case scenarios and provides planning-oriented Q\&A annotations together with trajectory data for training and evaluating autonomous driving models.

    \item \textbf{Alpamayo-R1}~\cite{wang2025alpamayo} introduces the Chain of Causation (CoC) annotations, featuring decision-grounded reasoning traces aligned with complex driving behaviors to enhance the interpretability and generalization of VLA models.
\end{itemize}

\paragraph{CoT Annotation Construction}
A key challenge in training \OneVL{} is obtaining high-quality chain-of-thought reasoning annotations paired with each driving scenario. On NAVSIM, we leverage the CoT annotations released by AdaThinkDrive~\cite{adathinkdrive}, the previous state-of-the-art method. These annotations provide natural-language reasoning traces that cover scene interpretation (such as lane boundaries), critical object analysis (including vehicles and pedestrians), and the final driving intent. They are synthesized by a VLM that converts raw detection labels (\eg, objects and lanes) into reasoning sequences, and serve as the supervision target for the language auxiliary decoder ($\mathcal{L}_{\text{l}}$). On ROADWork, we construct CoT annotations using a similar in-house pipeline. On Impromptu, we build CoT annotations from the original dataset's Q\&A pairs, augmenting the data with explicit decision and root-cause labels for corner-case trajectory prediction. On Alpamayo-R1, we use the released checkpoint to predict the CoC labels for all training examples. These annotations supervise the language auxiliary decoder, enabling the model to learn robust reasoning and decision-making logic in unstructured driving scenarios. Further details are provided in Appendix~\ref{sec:appendix:roadwork_cot}.

\subsection{Experimental Setups}
\label{sec:eval:setup}

\paragraph{Evaluation Metrics}
On \textbf{NAVSIM}, all methods are evaluated using the \textbf{Predictive Driver Model (PDM)} score, a composite metric that jointly assesses trajectory safety, comfort, and progress. On \textbf{ROADWork}, we report \textbf{ADE} (Average Displacement Error) and \textbf{FDE} (Final Displacement Error) to measure waypoint accuracy. On \textbf{Impromptu}, in addition to ADE and FDE, we also report the trajectory prediction \textbf{L2 error} over the first four seconds, following the protocol of the original paper. On \textbf{Alpamayo-R1}, we report ADE and FDE as well. Across all methods, we additionally report the average inference \textbf{latency}.

\paragraph{Baselines} We compare \OneVL{} against two categories of baselines, all built on Qwen3-VL-4B-Instruct~\cite{bai2025qwen3}, together with previous state-of-the-art methods as stronger reference points. 

The \emph{AR-based methods} that use standard autoregressive generation are:
\begin{itemize}
  \item \textbf{AR Answer}: Direct autoregressive trajectory prediction without any reasoning. The model receives the front-view image and ego state and directly outputs trajectory waypoints. This is the fastest baseline and defines the latency lower bound.
  
  \item \textbf{AR CoT+Answer}: Standard CoT reasoning followed by trajectory prediction. The model first generates a full reasoning chain and then produces the trajectory. This represents the performance upper bound for explicit reasoning, at the cost of substantially higher latency.
\end{itemize}

The \emph{Latent CoT methods} that use continuous latent representations for implicit reasoning are:
\begin{itemize}
  \item \textbf{COCONUT}~\cite{coconut}: Adapted for VLA-based autonomous driving. It uses curriculum learning to replace discrete reasoning tokens with continuous latent vectors.
  
  \item \textbf{CODI}~\cite{codi}: A COCONUT variant based on self-distillation, where a teacher model provides full textual CoT supervision and the student reasons in latent space.
  
  \item \textbf{SIM-CoT}~\cite{simcot}: A CODI variant that adds a separate text-decoding auxiliary decoder for language interpretability.
\end{itemize}

We also compare against previous \emph{state-of-the-art methods} reported in the literature. On NAVSIM (\emph{supervised fine-tuning setting}), these methods are:
\begin{itemize}
  \item \textbf{AdaThinkDrive}~\cite{adathinkdrive}: An 8B-parameter model with adaptive CoT reasoning for autonomous driving.
  
  \item \textbf{LaST-VLA}~\cite{lastvla}: An 8B-parameter vision-language-action model for autonomous driving.
\end{itemize}

On ROADWork, we compare against \textbf{YNet}~\cite{roadwork} and on Impromptu, we compare against the \textbf{Impromptu VLA}~\cite{chiimpromptu}. For Impromptu VLA, we report the result from our own replication. We also include a result obtained using the provided model checkpoint, which is shown in Table~\ref{tab:impromptu_reproduce_appendix}. On Alpamayo-R1, we compare against the \textbf{Cosmos-Reason}, a flow-matching-based VLA model. AR-based baselines are trained for 2 epochs with a learning rate of $4\times10^{-5}$ and batch size 64. Latent CoT baselines are trained for 6 epochs with a learning rate of $4\times10^{-5}$ and batch size 64. The results of previous state-of-the-art methods, where not explicitly stated otherwise, are taken directly from the literature.

\subsection{Main Results}
\label{sec:eval:main}

\begin{table*}[t]
\centering
\caption{\textbf{Performance comparisons on the NAVSIM benchmark.} PDM-score (higher is better) and average inference latency (lower is better) for all methods. Symbol $^{*}$ indicates the result is derived from the corresponding paper. For OneVL, we only count the parameters of the main VLM, as the auxiliary decoders are discarded during inference. The same applies to all subsequent models.}
\renewcommand{\arraystretch}{1.25}
\setlength{\tabcolsep}{9pt}
\begin{tabular}{lcccc}
\toprule
\textbf{Method} & \textbf{Model Size} & \textbf{PDM-score} $\uparrow$ & \textbf{Latency (s)} $\downarrow$ & \textbf{Interpretability} \\
\rowcolor[HTML]{FFE0CC}
\midrule
\multicolumn{5}{c}{\textit{Previous State-of-the-Art}} \\
$\circ$~AdaThinkDrive~\cite{adathinkdrive} & 8B & $86.20$$^{*}$ & -- & Language \\
$\circ$~LaST-VLA~\cite{lastvla} & 8B & $87.30$$^{*}$ & -- & -- \\
\midrule
\rowcolor[HTML]{FFE0CC}
\multicolumn{5}{c}{\textit{AR-based Baselines (4B, Qwen3-VL)}} \\
$\circ$~AR Answer & 4B & $87.47$ & \underline{$4.49$} & -- \\
$\circ$~AR CoT+Answer & 4B & \underline{$88.29$} & $6.58$ & Language \\
\midrule
\rowcolor[HTML]{FFE0CC}
\multicolumn{5}{c}{\textit{Latent CoT Baselines (4B, Qwen3-VL)}} \\
$\circ$~COCONUT~\cite{coconut} & 4B & $84.84$ & $5.93$ & -- \\
$\circ$~CODI~\cite{codi} & 4B & $83.92$ & $8.62$ & -- \\
$\circ$~SIM-CoT~\cite{simcot} & 4B & $84.21$ & $10.86$ & Language \\
\rowcolor[HTML]{FFE0CC}
$\bullet$~\textbf{OneVL}& 4B & $\mathbf{88.84}$ & $\mathbf{4.46}$ & Vision + Language \\
\bottomrule
\end{tabular}
\label{tab:main_results}
\vspace{0.1cm}
\end{table*}
\begin{table*}[!t]
\centering
\caption{\textbf{Performance comparisons on the ROADWork benchmark.} ADE and FDE (pixels; lower is better), latency (lower is better). Symbol $^{*}$ indicates the result is derived from the corresponding paper.}
\vspace{-0.1cm}
\renewcommand{\arraystretch}{1.25}
\setlength{\tabcolsep}{9pt}
\begin{tabular}{lcccc}
\toprule
\textbf{Method} & \textbf{ADE(pixel)} $\downarrow$ & \textbf{FDE(pixel)} $\downarrow$ & \textbf{Latency (s)} $\downarrow$ & \textbf{Interpretability} \\
\midrule
\rowcolor[HTML]{FFE0CC}
\multicolumn{5}{c}{\textit{Previous State-of-the-Art}} \\
$\circ$~YNet~\cite{roadwork} & $22.68$$^{*}$ & $80.78$$^{*}$ & -- & -- \\
\midrule
\rowcolor[HTML]{FFE0CC}
\multicolumn{5}{c}{\textit{AR-based Baselines (4B, Qwen3-VL)}} \\
$\circ$~AR Answer            & $15.98$ & $40.29$ & \underline{$4.74$} & -- \\
$\circ$~AR CoT+Answer        & \underline{$13.18$} & \underline{$29.98$} & $10.74$ & Language \\
\midrule
\rowcolor[HTML]{FFE0CC}
\multicolumn{5}{c}{\textit{Latent CoT Baselines (4B, Qwen3-VL)}} \\
$\circ$~COCONUT~\cite{coconut}              & $15.44$ & $38.60$ & \underline{$6.06$} & -- \\
$\circ$~CODI~\cite{codi}                 & $16.45$ & $44.28$ & $6.73$ & -- \\
$\circ$~SIM-CoT~\cite{simcot}              & $16.49$ & $44.32$ & $6.19$ & Language \\
\rowcolor[HTML]{FFE0CC}
$\bullet$~\textbf{OneVL}           & $\mathbf{12.49}$ & $\mathbf{28.80}$ & $\mathbf{4.71}$ & Vision + Language \\
\bottomrule
\end{tabular}
\label{tab:roadwork_results}
\vspace{0.2cm}
\end{table*}
 
\begin{table*}[!t]
\centering
\caption{\textbf{Performance comparisons on the Impromptu benchmark.} ADE and FDE (meters; lower is better), latency (lower is better).}
\vspace{-0.1cm}
\renewcommand{\arraystretch}{1.25}
\setlength{\tabcolsep}{9pt}
\begin{tabular}{lccccc}
\toprule
\textbf{Method}& \textbf{Model Size} & \textbf{ADE(m)} $\downarrow$ & \textbf{FDE(m)} $\downarrow$ & \textbf{Latency (s)} $\downarrow$ & \textbf{Interpretability} \\
\midrule
\rowcolor[HTML]{FFE0CC}
\multicolumn{6}{c}{\textit{Previous State-of-the-Art}} \\
$\circ$~Impromptu VLA~\cite{chiimpromptu} & 3B & $1.60$ & $4.28$  & $6.10$ & -- \\
\midrule
\rowcolor[HTML]{FFE0CC}
\multicolumn{6}{c}{\textit{AR-based Baselines (4B, Qwen3-VL)}} \\
$\circ$~AR Answer       & 4B     & $1.46$  & $4.03$ & \underline{$4.24$} & -- \\
$\circ$~AR CoT+Answer  & 4B      &\underline{$1.42$} &\underline{$3.96$} & $6.84$ & Language \\
\midrule
\rowcolor[HTML]{FFE0CC}
\multicolumn{6}{c}{\textit{Latent CoT Baselines (4B, Qwen3-VL)}} \\
$\circ$~COCONUT~\cite{coconut}   & 4B           & $1.49$ & $4.07$ & $5.27$ & -- \\
$\circ$~CODI~\cite{codi}   & 4B              & $1.86$ & $5.18$ & $5.24$ & -- \\
$\circ$~SIM-CoT~\cite{simcot}   & 4B           & $2.43$ & $6.10$ & $5.09$ & Language \\
\rowcolor[HTML]{FFE0CC}
$\bullet$~\textbf{OneVL}     & 4B      & $\mathbf{1.34}$  & $\mathbf{3.70}$  & $\mathbf{4.02}$  & Vision + Language \\
\bottomrule
\end{tabular}
\label{tab:impromptu_results}
\vspace{0.2cm}
\end{table*}
\begin{table*}[!t]
\centering
\caption{\textbf{Performance comparisons on the Impromptu benchmark.} We report trajectory prediction L2 error following the benchmark setting (meters; lower is better).}
\vspace{-0.1cm}
\renewcommand{\arraystretch}{1.25}
\setlength{\tabcolsep}{17pt}
\begin{tabular}{lcccccc}
\toprule
\multicolumn{6}{c}{\textbf{Traj. Pred. L2 Error (m)}} \\
\textbf{Method} & \textbf{1s} $\downarrow$ & \textbf{2s} $\downarrow$ & \textbf{3s} $\downarrow$ & \textbf{4s} $\downarrow$  & \textbf{Avg.} $\downarrow$ \\
\midrule
\rowcolor[HTML]{FFE0CC}
\multicolumn{6}{c}{\textit{Previous State-of-the-Art}} \\
$\circ$~Impromptu VLA~\cite{chiimpromptu} & $0.14$ & $0.60$  & $1.45$ & $2.67$ & $1.22$ \\
\midrule
\rowcolor[HTML]{FFE0CC}
\multicolumn{6}{c}{\textit{AR-based Baselines (4B, Qwen3-VL)}} \\
$\circ$~AR Answer            & \underline{$0.13$} & \underline{$0.51$} & $1.29$ & $2.46$ & $1.11$ \\
$\circ$~AR CoT+Answer        & \underline{$0.13$} & \underline{$0.51$} & \underline{$1.27$} & \underline{$2.44$} & \underline{$1.09$} \\
\midrule
\rowcolor[HTML]{FFE0CC}
\multicolumn{6}{c}{\textit{Latent CoT Baselines (4B, Qwen3-VL)}} \\
$\circ$~COCONUT~\cite{coconut}              & $0.15$ & $0.54$ & $1.32$ & $2.50$ & $1.13$ \\
$\circ$~CODI~\cite{codi}                 & $0.17$ & $0.63$ & $1.61$ & $3.13$ & $1.39$\\
$\circ$~SIM-CoT~\cite{simcot}              & $0.41$ & $1.10$ & $2.25$ & $3.94$ & $1.93$ \\
\rowcolor[HTML]{FFE0CC}
$\bullet$~\textbf{OneVL}           &  $\mathbf{0.13}$ & $\mathbf{0.48}$ & $\mathbf{1.18}$ & $\mathbf{2.25}$ & $\mathbf{1.01}$ \\
\bottomrule
\end{tabular}
\label{tab:impromptu_results_appendix}
\vspace{0.2cm}
\end{table*}

\begin{table*}[!t]
\centering
\caption{\textbf{Performance comparisons on the Alpamayo-R1 benchmark.} ADE and FDE (meters; lower is better), latency (lower is better).}
\vspace{-0.1cm}
\renewcommand{\arraystretch}{1.25}
\setlength{\tabcolsep}{9pt}
\begin{tabular}{lccccc}
\toprule
\textbf{Method} & \textbf{Model Size} &\textbf{ADE(m)} $\downarrow$ & \textbf{FDE(m)} $\downarrow$ & \textbf{Latency (s)} $\downarrow$ & \textbf{Interpretability} \\
\midrule
\rowcolor[HTML]{FFE0CC}
\multicolumn{6}{c}{\textit{Previous State-of-the-Art}} \\
$\circ$~Cosmos-Reason & 10B & \underline{$2.86$} &  $\mathbf{7.42}$ & -- & Language \\
\midrule
\rowcolor[HTML]{FFE0CC}
\multicolumn{6}{c}{\textit{AR-based Baselines (4B, Qwen3-VL)}} \\
$\circ$~AR Answer       & 4B     &  $3.27$ & $9.59$ & $\mathbf{3.06}$ & -- \\
$\circ$~AR CoT+Answer     & 4B    & $2.99$ & $8.54$ & $3.51$ & Language \\
\midrule
\rowcolor[HTML]{FFE0CC}
\multicolumn{6}{c}{\textit{Latent CoT Baselines (4B, Qwen3-VL)}} \\
$\circ$~COCONUT~\cite{coconut}  & 4B             &  $3.29$ & $9.48$ & $3.76$ & -- \\
$\circ$~CODI~\cite{codi}       & 4B           & $3.22$ & $9.25$ & $3.85$ & -- \\
$\circ$~SIM-CoT~\cite{simcot}    & 4B           & $3.40$ & $9.85$ & $3.78$ & Language \\
\rowcolor[HTML]{FFE0CC}
$\bullet$~\textbf{OneVL}     & 4B       &  $\mathbf{2.62}$ &  \underline{$7.53$} &  \underline{$3.23$} & Vision + Language \\
\bottomrule
\end{tabular}
\label{tab:ar1_results}
\vspace{0.2cm}
\end{table*}

Table~\ref{tab:main_results} presents the NAVSIM comparison results. Table~\ref{tab:roadwork_results} reports ADE/FDE on ROADWork. Table~\ref{tab:impromptu_results} and Table~\ref{tab:impromptu_results_appendix} report the score on Impromptu. Table~\ref{tab:ar1_results} reports the overall results on Alpamayo-R1. Figure~\ref{fig:teaser} provides a visual overview of all the results. We can make the following observations from the results. 

\paragraph{OneVL achieves best performance}
Specifically, \OneVL{} achieves 88.84 PDM-score, surpassing the previous supervised finetuned SOTA AdaThinkDrive (86.20) and LaST-VLA (87.30) by +2.64 and +1.54, respectively. On ROADWork, \OneVL{} also achieves 12.49 ADE and 28.80 FDE, significantly surpassing the previous SOTA YNet (22.68/80.78), respectively. The same pattern is observed on the Impromptu and Alpamayo-R1 datasets. Besides, \OneVL{} outperforms all AR-based, latent CoT, and notably, explicit CoT baselines. This consistent superiority demonstrates the effectiveness of our multimodal auxiliary supervision approach in advancing planning tasks. On the Alpamayo-R1 dataset, OneVL underperforms in FDE with a slight margin (7.53 \emph{vs.} 7.42). This is expected, as Cosmos-Reason uses RL to further enhance its capabilities.

\paragraph{Prefill inference matches answer-only prediction speed}
On NAVSIM, \OneVL{} with prefill inference achieves 4.46 latency, essentially identical to AR answer-only prediction (4.49s). This validates the core efficiency claim of our approach. By prefilling latent tokens into the prompt context (which is processed in a single parallel prefill pass), we incur negligible additional latency compared to sequential autoregressive generation of explicit reasoning tokens. On ROADWork, Impromptu, and Alpamayo-R1, we observe a similar speed-up effect, where \OneVL{} achieves 4.71s, 4.02s, and 3.23s latency respectively, which are comparable to or even lower than those of answer-only prediction. In all cases, \OneVL{} inference is much faster than the AR CoT+Answer baseline.

\paragraph{Existing latent CoT methods fail on VLA-based autonomous driving}
All three adapted latent CoT methods perform substantially worse than answer-only AR prediction, except for the Alpamayo-R1 dataset. This is a critical finding that purely linguistic latent CoT approaches that were designed for text-only reasoning tasks do not transfer effectively to the multimodal spatial-temporal domain of autonomous driving trajectory prediction. The compact linguistic latent spaces these methods define are insufficient to support the geometric reasoning required for precise waypoint prediction. From the compression view of intelligence, this failure is fundamental rather than incidental. Concretely, these methods compress language descriptions of the scene, but language is already an abstraction of the physical world—it encodes semantic labels and relationships, not the spatial-temporal causal dynamics that determine future outcomes. In other words, it satisfies the efficiency criterion of the compression principle but not the intelligence criterion.
A further likely factor is that these baselines are not trained with \OneVL{}'s staged recipe (\Cref{sec:training}), so their latent streams may remain poorly aligned with trajectory and multimodal supervision when optimization starts fully coupled.

\OneVL{} overcomes this limitation by pairing compression with verification: the auxiliary decoders ensure that the latent bottleneck preserves spatially and linguistically meaningful content, rather than collapsing to degenerate representations. This is why \OneVL{} is the only latent CoT method that outperforms the AR baseline. The three-stage training recipe is equally crucial, as it progressively aligns the bottleneck before joint optimization. More analysis can be found in~\Cref{sec:eval:analysis}.

\paragraph{Explicit CoT supervision helps: AR CoT+Answer \emph{vs.} AR Answer}
Comparing AR CoT+Answer (88.29) to AR Answer (87.47) on NAVSIM confirms that explicit reasoning supervision provides meaningful trajectory improvements (+0.80 PDM-score). We can observe similar patterns on ROADWork, Impromptu, and Alpamayo-R1. These results confirm the core motivation for incorporating CoT reasoning into autonomous driving systems and establish the quality ceiling that latent CoT methods aim to approach while reducing latency.

\subsection{Explanation Quality}
\label{sec:eval:explanation}

\begin{figure}[!t]
    \centering
    \includegraphics[width=\textwidth]{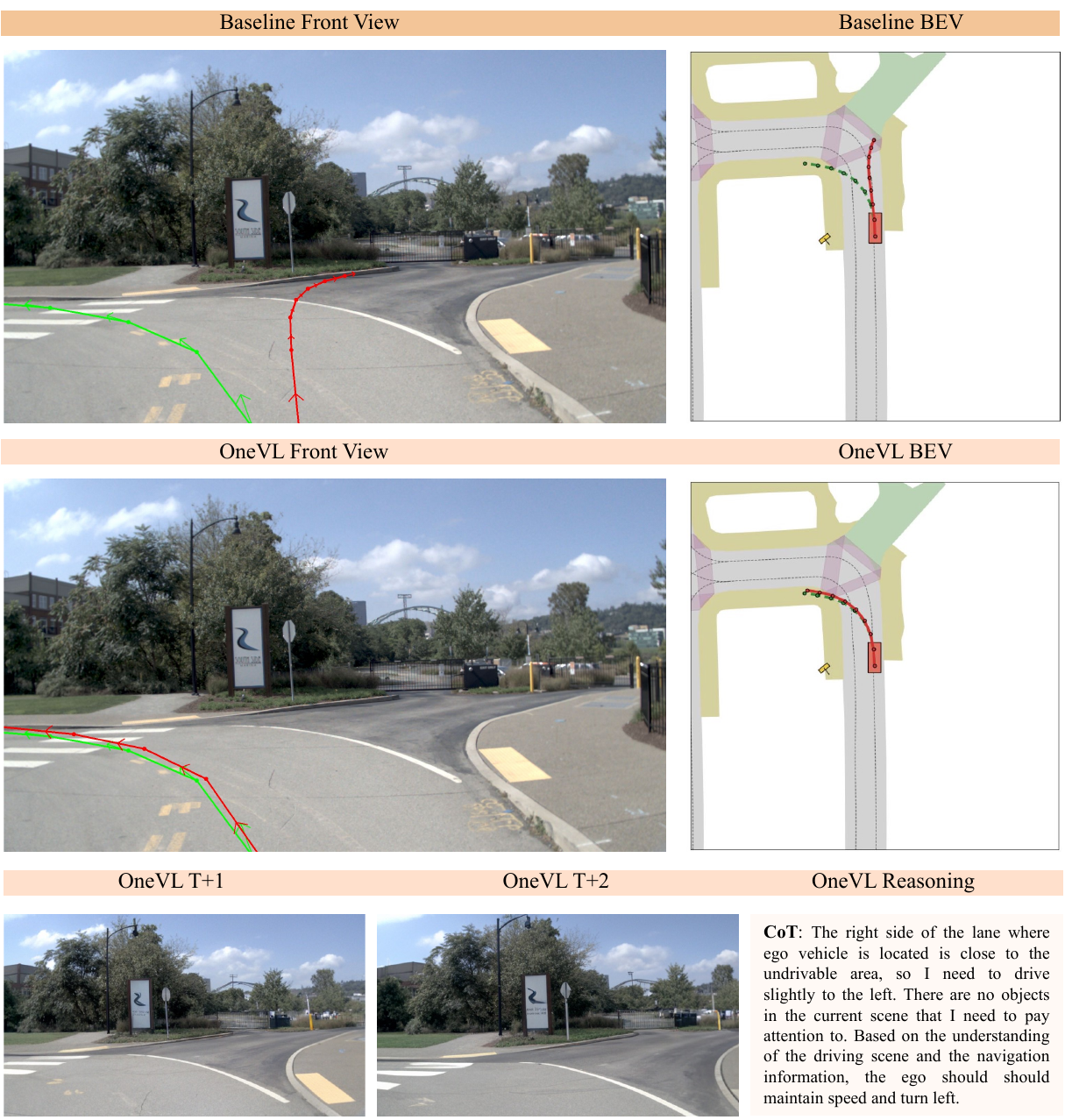} 
   \caption{\textbf{Visualizations of prediction on NAVSIM.}  Each plot overlays ground-truth (\textcolor{green!70!black}{green}) and predicted (\textcolor{red!70!black}{red}) trajectories on the front camera view. More examples can be found in Appendix~\ref{sec:appendix:navsim_qualitative}.}
\label{fig:qualitative_navsim_main}
\end{figure}

\begin{figure}[!t]
    \centering
    \includegraphics[width=\textwidth]{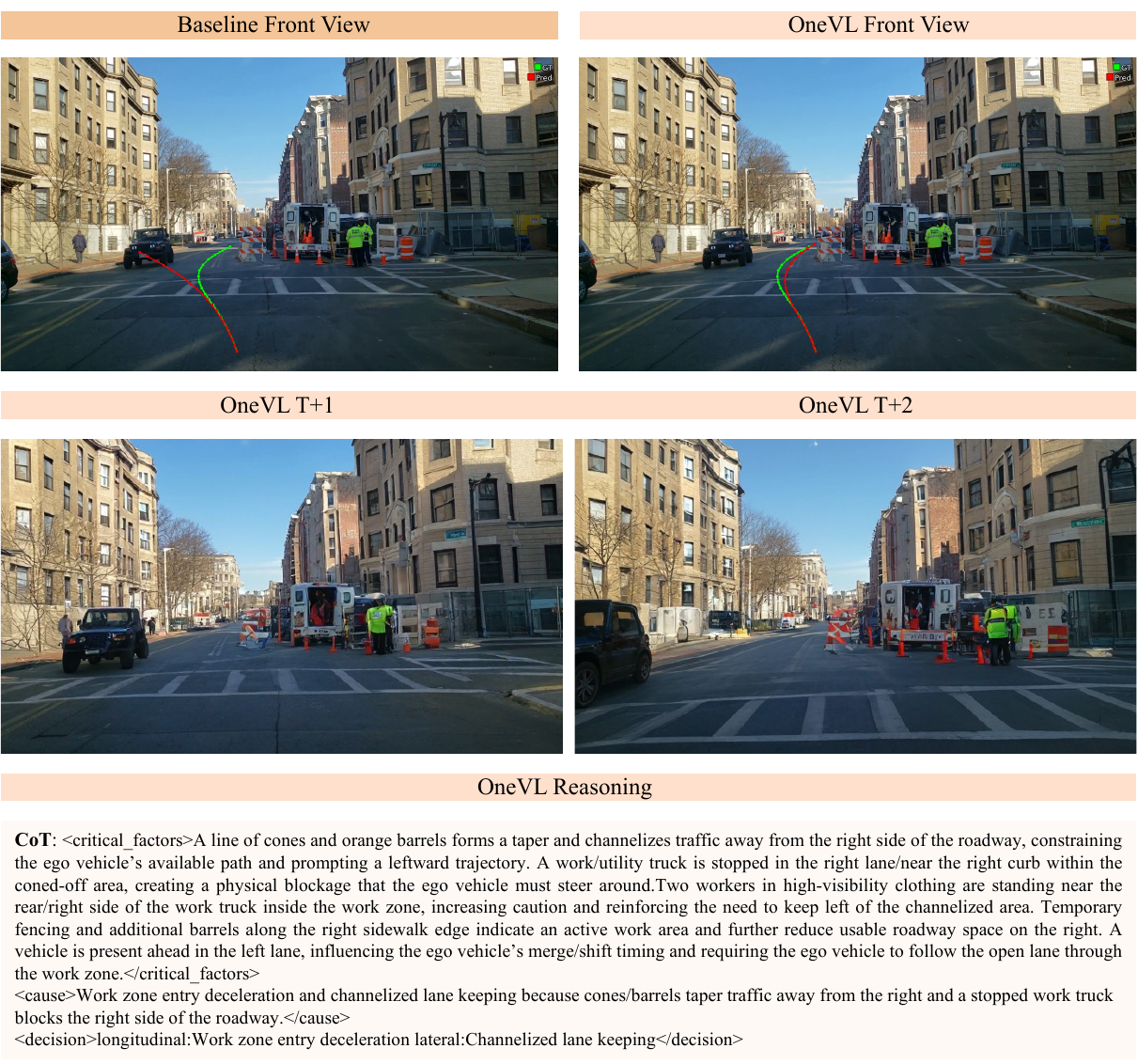} 
   \caption{\textbf{Visualizations of prediction on ROADWork.}  Each plot overlays ground-truth (\textcolor{green!70!black}{green}) and predicted (\textcolor{red!70!black}{red}) trajectories on the front camera view. More examples can be found in Appendix~\ref{sec:appendix:roadwork_qualitative}.}
\label{fig:qualitative_roadwork_main}
\end{figure}

\begin{figure}[!t]
    \centering
    \includegraphics[width=\textwidth]{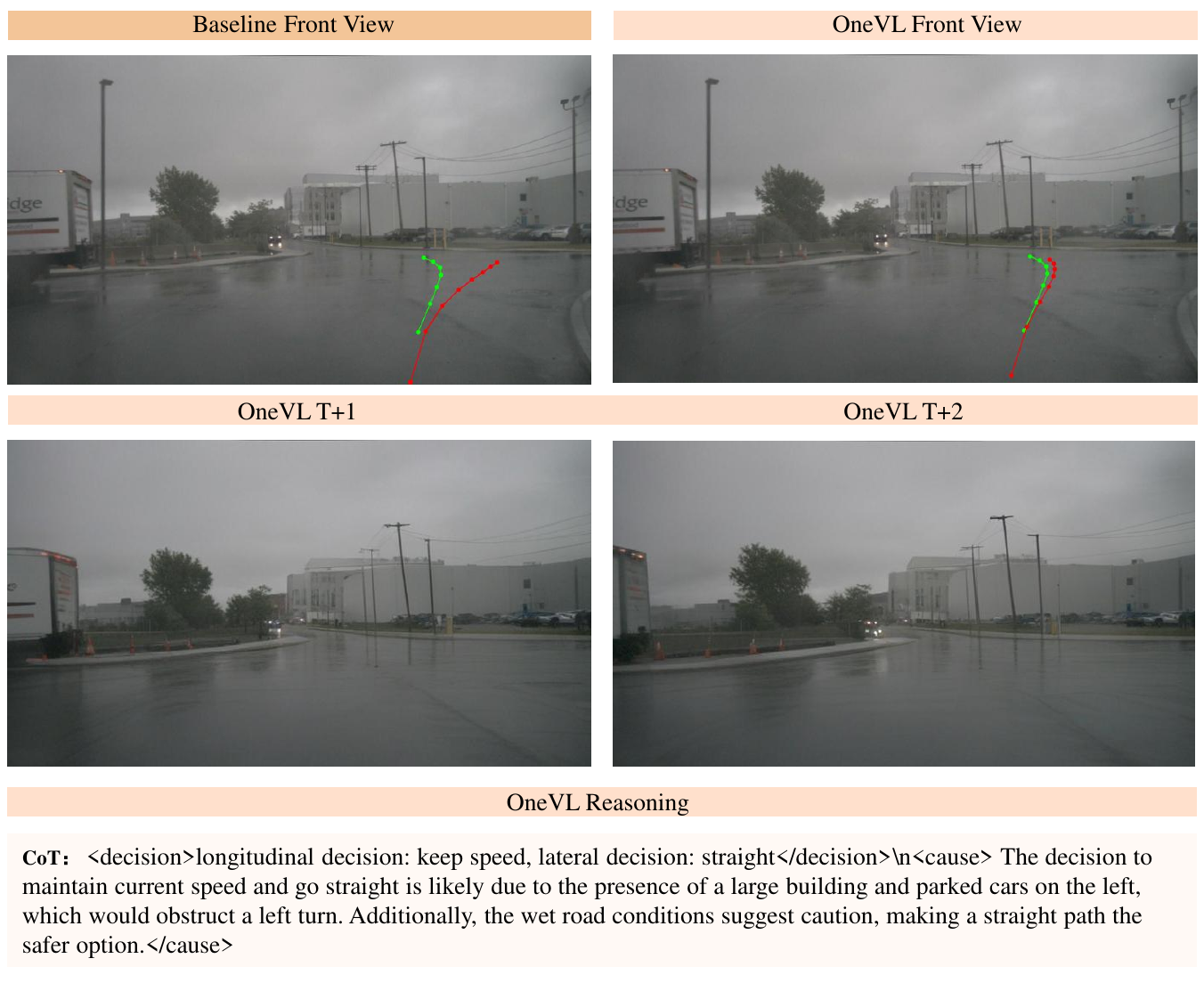} 
   \caption{\textbf{Visualizations of prediction on Impromptu.}  Each plot overlays ground-truth (\textcolor{green!70!black}{green}) and predicted (\textcolor{red!70!black}{red}) trajectories on the front camera view. More examples can be found in Appendix~\ref{sec:appendix:impromptu_qualitative}.}
\label{fig:qualitative_impromptu_main}
\end{figure}

\begin{figure}[!t]
    \centering
    \includegraphics[width=\textwidth]{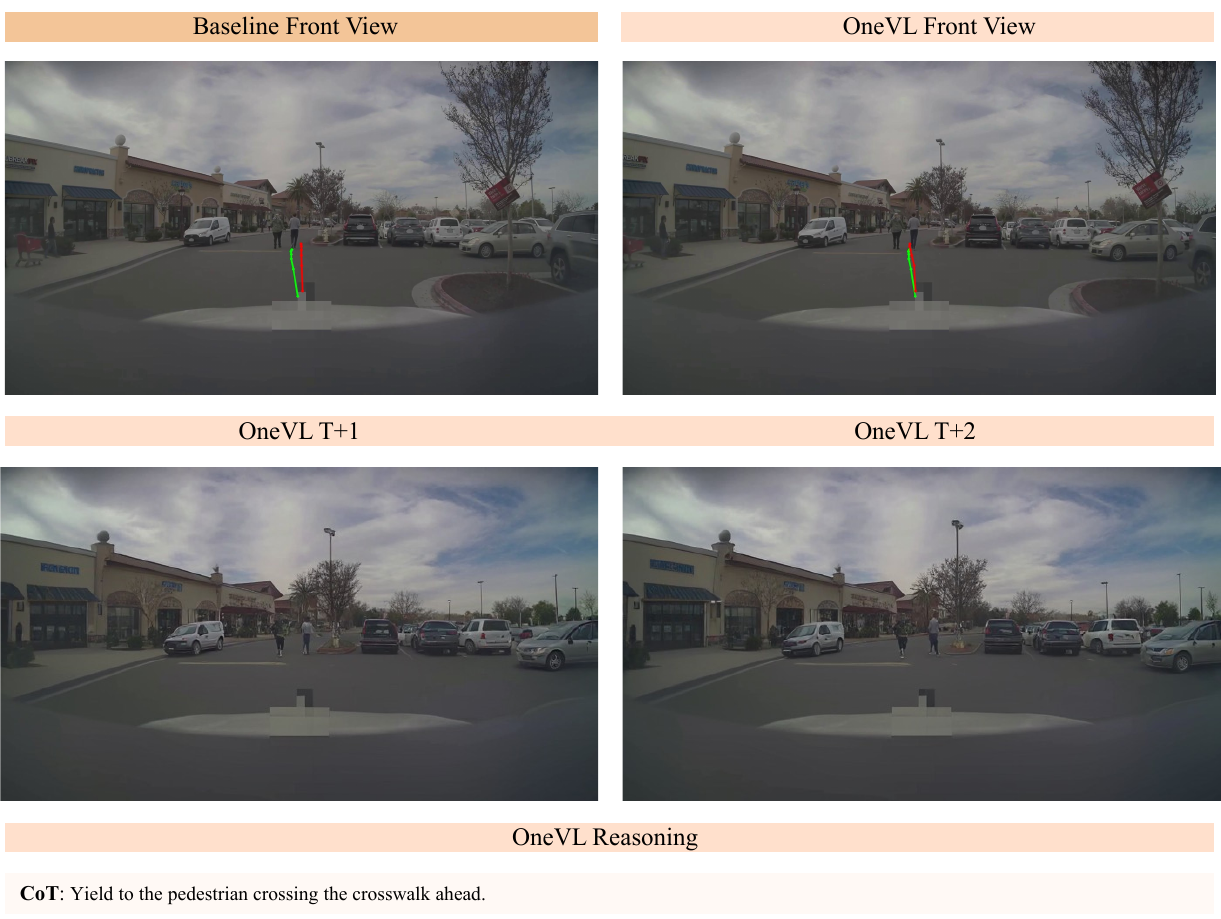} 
   \caption{\textbf{Visualizations of prediction on Alpamayo-R1.}  Each plot overlays ground-truth (\textcolor{green!70!black}{green}) and predicted (\textcolor{red!70!black}{red}) trajectories on the front camera view. More examples can be found in Appendix~\ref{sec:appendix:apr1_qualitative}.}
\label{fig:qualitative_ar1}
\end{figure}

Beyond trajectory prediction accuracy, \OneVL{} is unique in providing human-interpretable explanations in both language and vision. Figures~\ref{fig:qualitative_navsim_main}, \ref{fig:qualitative_roadwork_main}, \ref{fig:qualitative_impromptu_main} and~\ref{fig:qualitative_ar1} provide NAVSIM, ROADWork, Impromptu and Alpamayo-R1 qualitative examples, including a side-by-side trajectory prediction comparison (\emph{vs.} AR Answer baseline), two future frames from the visual auxiliary decoder, and the text CoT from the language auxiliary decoder. We also quantitatively evaluate the quality of text explanations.

\paragraph{Text CoT Quality}
\label{sec:eval:text_cot}
We evaluate the language explanations produced by the language auxiliary decoder against the ground-truth CoT annotations on the 500 NAVSIM test set. We compare against the AR CoT+Answer baseline (which generates explicit CoT autoregressively) and SIM-CoT (the only prior latent CoT method with language interpretability). We report three complementary metrics that capture different dimensions of explanation quality:

\begin{itemize}
  \item \textbf{Meta Action Accuracy}: Following~\citet{luo2025mtrdrive}, we report the meta action accuracy. On NAVSIM, each CoT concludes with a high-level driving decision (\eg, ``the ego should maintain speed and keep lane''). We extract this meta-action clause from both the ground-truth and the predicted CoT and compute exact string match accuracy. This metric directly measures whether the model's reasoning arrives at the correct driving intent, which is the most safety-critical aspect of CoT quality.
  \item \textbf{STS Score (Semantic Textual Similarity Score)}: We compute a neural semantic similarity score between each predicted CoT and its ground-truth reference using a cross-encoder reranker (BGE-reranker-v2-m3~\cite{chen2024bge}). This evaluator is particularly suitable for computing a similarity score for templated CoTs where most of the words are identical. A cross-encoder concatenates the ground truth and the prediction, processing them simultaneously through full token-by-token cross-attention. This mechanism allows the model to perform a deep, comparative analysis, making it highly sensitive to critical localized contradictions—such as predicting ``slowly'' instead of ``fastly'', or outputting hazardous numerical distances. For each ground-truth-prediction pair, the cross-encoder outputs a raw relevance logit. We apply global min-max normalization across all methods and examples to obtain scores in $[0, 1]$. This metric captures fine-grained semantic alignment beyond surface-level n-gram overlap.
  
  \item \textbf{LLM-as-Judge Score}: Following~\citet{luo2025mtrdrive,ishaq2025drivelmm}, we also employ a state-of-the-art proprietary VLM, gemini-3.1-flash-lite-preview, as an automated evaluator. Given the front-camera image, the ground-truth CoT, and the predicted CoT, the judge model scores each prediction on a 0-100 scale based on four criteria: (1) perception accuracy; (2) motion state prediction; (3) ego decision correctness, and (4) language fluency. Scores are normalized to $[0, 1]$. This metric provides a holistic evaluation that accounts for both visual grounding and reasoning quality. The full evaluation prompt is provided in Appendix~\ref{sec:appendix:llm_judge_prompt}.
\end{itemize}

Quantitative results are shown in Table~\ref{tab:text_cot_quality}. OneVL consistently outperforms SIM-CoT across all evaluation dimensions. Specifically, OneVL achieves a Meta Action Accuracy of 71.00, a significant 3.8 improvement over SIM-CoT (67.20). This indicates that our latent representation more effectively captures and decodes the critical high-level driving intents. In terms of semantic alignment, OneVL reaches an STS score of 78.26 and an LLM Judge score of 79.13, narrowing the gap to the autoregressive baseline (AR CoT+Answer). While the AR model maintains the highest scores across all metrics, it requires explicit sequential text generation. This results in higher inference latency as demonstrated in Table~\ref{tab:main_results}. In contrast, OneVL achieves competitive explanation quality while benefiting from the efficiency of a latent reasoning framework. These results demonstrate that the language auxiliary decoder in OneVL successfully enables faithful, human-readable explanations of the model's reasoning.

\begin{table}[t]
\centering
\caption{\textbf{Text CoT quality on the NAVSIM test set.}
  \textit{Meta Action Acc.}\ measures exact match of the predicted driving decision against the ground-truth meta action. \textit{STS score} reports the mean semantic similarity score (global min-max normalized). \textit{LLM Judge} reports the mean VLM evaluation score (normalized to $[0, 1]$; see Appendix~\ref{sec:appendix:llm_judge_prompt} for the full prompt). Higher is better for all metrics.}
\vspace{-0.1cm}
\renewcommand{\arraystretch}{1.3}
\setlength{\tabcolsep}{11pt}
\begin{tabular}{lcccc}
\toprule
\textbf{Method} & \textbf{Meta Action Acc.} $\uparrow$ & \textbf{STS} $\uparrow$ & \textbf{LLM Judge} $\uparrow$ & \textbf{Avg.} $\uparrow$ \\
\midrule
$\circ$~AR CoT+Answer        & \textbf{73.20} & \textbf{79.75} & \textbf{81.86} & \textbf{78.27} \\
$\circ$~SIM-CoT              & 67.20 & 76.25 & 78.73 & 74.06\\
\rowcolor[HTML]{FFE0CC}
$\bullet$~\textbf{\OneVL{}} (lang.\ aux.) & \underline{71.00} &  \underline{78.26} &  \underline{79.13} &  \underline{76.13} \\
\bottomrule
\end{tabular}
\label{tab:text_cot_quality}
\vspace{0.2cm}
\end{table}

\subsection{Ablation Study}
\label{sec:eval:ablation}

\begin{table*}[t]
\centering
\caption{\textbf{Ablation study on each component in OneVL}.}
\vspace{-0.1cm}
\renewcommand{\arraystretch}{1.25}
\setlength{\tabcolsep}{4pt}
\begin{tabular}{lccccc}
\toprule
\textbf{Model Variant} & \textbf{Lang.\ Aux.\ Dec.} & \textbf{Vis.\ Aux.\ Dec.} & \textbf{Staged Train} & \textbf{PDM-score} $\uparrow$ \\
\midrule
$\circ$~\OneVL{} w/o vis. dec. & \Checkmark & -- & \Checkmark  &87.97 \\
$\circ$~\OneVL{} w/o lang.e dec. & -- & \Checkmark & \Checkmark  & \underline{88.53} \\
$\circ$~\OneVL{} w/o staged train & \Checkmark & \Checkmark & --  &67.13 \\
\rowcolor[HTML]{FFE0CC}
$\bullet$~\textbf{\OneVL{}} (full configuration) & \Checkmark & \Checkmark &  \Checkmark  &\textbf{88.84} \\
\bottomrule
\end{tabular}
\label{tab:ablation}
\vspace{0.2cm}
\end{table*}

To understand the contribution of each component, we conduct ablation experiments by training additional models with different components. OneVL w/o language decoder, OneVL w/o visual decoder, and OneVL w/o staged train are trained. For OneVL w/o language decoder, and OneVL w/o visual decoder, we follow the same training recipe as \OneVL{}, but without the language or visual auxiliary decoders. For OneVL w/o staged train, we use the same architecture as \OneVL{}, but directly optimize the model with the end-to-end joint learning with the same training setting as stage 3. Results are shown in Table~\ref{tab:ablation}.

Comparing \OneVL{} w/o visual decoder (87.97) to the full \OneVL{} model (88.84), we find that the visual auxiliary decoder contributes +0.87 score. This result confirms that visual latent supervision—requiring the model to encode anticipated future scene content—provides complementary and additive benefit beyond language reasoning alone.
A comparison between \OneVL{} without its language decoder (88.53) and the full \OneVL{} model (88.84) validates the contribution of the language auxiliary decoder and language latent tokens, yielding a modest performance gain of +0.31. While the absolute improvement is small, it demonstrates that encoding language-based reasoning into compact latent tokens can enhance trajectory prediction performance.

\begin{figure}[!t]
\vspace{0.1cm}
    \centering
    \includegraphics[width=\textwidth]{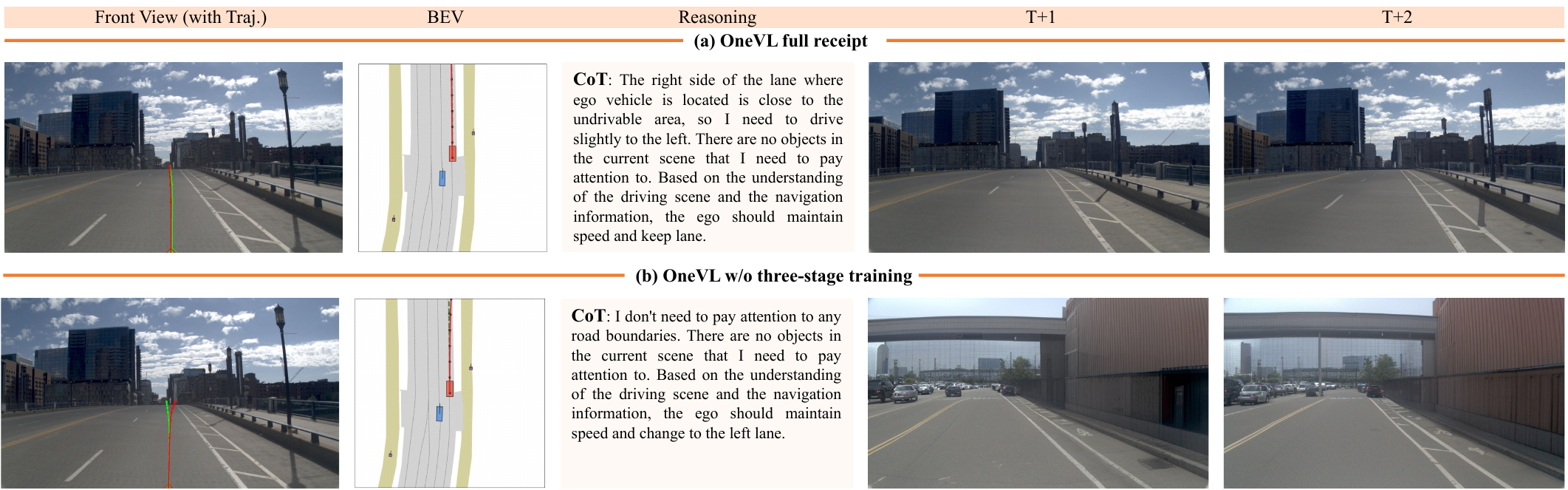} 
    \caption{\textbf{Visual CoT under full training \emph{vs.}\ an overfitting visual auxiliary decoder.} (a) Top row: with the full training recipe, latent visual tokens decode to future frames that remain scene-consistent and usable as spatial-temporal reasoning supervision. (b) Bottom row: when without three-stage training, decoding from the same input collapses to memorized artifacts rather than generalizing---the predicted ``future'' frames are visually irrelevant to the input image.}
\label{fig:ablation_visual_cot}
\vspace{0.1cm}
\end{figure}

\paragraph{On the efficacy of three-stage training}
The ablation study comparing OneVL without three-stage training to the full model yields a conclusive result: the proposed three-stage training strategy is not merely beneficial but essential. Direct end-to-end joint fine-tuning fails catastrophically, causing the PDM-Score to drop by 21.71 points (from 88.84 to 67.13). This drastic performance gap underscores the necessity of the staged approach.

A detailed examination of the training dynamics reveals the underlying causes. First, the direct approach suffers from severe ``gradient shock'' at initialization, with an exploding gradient norm of 378.22 that destabilizes the pre-trained backbone. In contrast, the three-stage strategy maintains a stable gradient norm of 0.28 by properly warming up the latent tokens. Second, the end-to-end method causes catastrophic task interference. The backbone struggles to optimize conflicting objectives simultaneously, resulting in a much higher final trajectory prediction loss (0.186 \emph{vs.} 0.136), showing that the model converges into a local minimum and consequently has poor driving performance.

We further examine the visual quality of predicted future frames. Figure~\ref{fig:ablation_visual_cot} compares decoded future frames from the same NAVSIM test image. Under the full OneVL recipe (a), the visual auxiliary decoder produces spatially coherent previews for $t+1$ and $t+2$ that match plausible ego motion and scene layout, demonstrating a consistent and reasonable visual chain-of-thought. In contrast, for the ablated training setup (b), the decoded frames are totally irrelevant to the input, which introduces noise. 

Besides, the language CoT reasoning in (b) is also erroneous. This demonstrates that without the staged training, the model memorized training patterns rather than learning real-world dynamics. We also observe that the full recipe's visual explain loss (\ie, $\mathcal{L}_v$) curve is smooth and stable across training, whereas the ablation curve exhibits a pronounced spike at the beginning. All observations indicate that the visual decoder fails to learn reliable scene dynamics without staged training. This indicates that without the three-stage curriculum, the visual auxiliary decoder overfits, taking shortcuts and memorizing data instead of learning to generalize. 

Consequently, the visual supervision signal fails to improve the final trajectory prediction and instead introduces noise that degrades overall performance. This observation underscores that effective compression requires careful optimization: the three-stage curriculum ensures the bottleneck learns to encode generalizable scene dynamics rather than memorized shortcuts, a concrete instance of how the quality of compression determines the quality of intelligence.

\subsection{Towards Real-World Deployment}
\label{sec:eval:deployment}

Closed-loop on-vehicle testing favors extremely low-latency trajectory prediction. However, autoregressive decoding waypoints still dominate the budget even when latent reasoning is prefilled. To explore a deployment-oriented variant, we append a compact MLP head on top of the same Qwen3-VL-4B-Instruct backbone (using the hidden state of the last latent token as the input). Therefore, the model can autoregressively predict the trajectory using the LLM head as well as predict the trajectory with a single feed-forward pass using the MLP head. This variant can still exploit multimodal latent supervision during training.

Table~\ref{tab:deployment_mlp} compares this MLP variant against full \OneVL{} on NAVSIM. The MLP head reaches 86.83 PDM-score with 0.24s inference latency, whereas full \OneVL{} keeps 88.84 with 4.46s inference latency, achieving a 5.4\% of the AR model’s latency. At $1/0.2405 \approx 4.16$\,Hz end-to-end, the MLP variant meets a typical on-vehicle few-Hz budget while remaining in a competitive performance relative to strong prior models (\eg, LaST-VLA). This promising outcome supports the potential for real-world vehicle deployment.

\begin{table}[!t]
\centering
\caption{\textbf{Accuracy-latency trade-off for real-time deployment on NAVSIM.}}
\vspace{-0.1cm}
\renewcommand{\arraystretch}{1.2}
\setlength{\tabcolsep}{8pt}
\begin{tabular}{lcc}
\toprule
\textbf{Variant} & \textbf{PDM-score} $\uparrow$ & \textbf{latency (s)} $\downarrow$ \\
\midrule
$\circ$~OneVL (regression) & 86.83 & \textbf{0.24} \\
\rowcolor[HTML]{FFE0CC}
$\bullet$~\textbf{\OneVL{}} (AR) & \textbf{88.84}  & 4.46 \\
\bottomrule
\end{tabular}
\label{tab:deployment_mlp}
\vspace{0.2cm}
\end{table}

\subsection{In-Depth Analysis: Where Does the Benefit Come From?}
\label{sec:eval:analysis}

\paragraph{Implicit Reasoning \emph{vs.} Explicit Tokens}
A natural question is why \OneVL{}'s implicit latent reasoning achieves better performance than explicit AR CoT+Answer (88.84 \emph{vs.} 88.29), despite using fewer tokens to express the reasoning. We hypothesize two mechanisms:

First, the \emph{compression benefit}: compact latent tokens force the model to distill the most trajectory-relevant reasoning into a small representational bottleneck, filtering out irrelevant or redundant content. This is precisely the mechanism predicted by the information bottleneck principle~\cite{tishby1999information}. Tighter compression discards noise and retains only the causal features that are predictive of the output, yielding representations that generalize better than verbose, free-form CoT chains, where tangential reasoning may introduce noise into the trajectory prediction.

Second, the \emph{world model grounding benefit}: the visual auxiliary decoder objective explicitly requires the visual latent tokens to encode spatial-temporal scene dynamics (future frame content), which is directly relevant to trajectory prediction. This is a world model supervision signal—predicting what the scene will look like forces the compressed latents to internalize the causal dynamics of agent motion and road geometry. Explicit language CoT does not have an analogous spatial grounding mechanism; it describes the world symbolically, leaving the causal geometry implicit.

\paragraph{Why Visual Supervision Helps More Than Language Supervision}
The visual auxiliary decoder contributes +0.87 PDM-score compared to +0.31 for the language auxiliary decoder. This asymmetry reflects the world model role the visual decoder plays. Autonomous driving trajectory prediction is fundamentally a spatial prediction task, and visual token reconstruction, \ie, predicting what the scene looks like 0.5–1.0 seconds later, provides a supervision signal that is inherently aligned with the geometric nature of trajectory prediction. 

Crucially, future-frame prediction is a world model objective: to minimize reconstruction loss on unseen configurations of agents and road geometry, the visual latent tokens must encode the causal dynamics of the scene, not just its current appearance. Language CoT annotation describes the reasoning process in abstract, symbolic terms; it is valuable for semantic grounding but is one step removed from the physical dynamics that drive trajectory outcomes. The world model decoder thus provides a harder, more causally direct compression target that language supervision alone cannot supply, which is precisely why it contributes a larger performance gain.

\paragraph{Latency Analysis}
The prefill inference mechanism achieves its speed advantage because modern transformer implementations process the prefill sequence in a single parallel forward pass, while autoregressive decoding requires sequential token generation. With several additional latent tokens in the prefill phase, the overhead is negligible compared to the much longer image patch sequence already in the prefill context. For example, on NAVSIM, the result is that prefill-mode \OneVL{} (4.46s) is essentially indistinguishable from AR Answer (4.49s) in terms of latency, matching the speed of answer-only autoregressive prediction. Further experiment in~\Cref{sec:eval:deployment} demonstrates that by introducing the MLP head, OneVL can achieve the optimal inference speed (0.24s) at the cost of performance.

\paragraph{Why Prior Latent CoT Methods Fail on Autonomous Driving}
The catastrophic failure of COCONUT, CODI, and SIM-CoT warrants deeper analysis. These methods were designed under the assumption that latent representations can compress reasoning quality from teacher text generation into student latent tokens via semantic alignment. However, in the autonomous driving domain, these methods fail because:

\begin{itemize}
    \item \textbf{Lack of visual world model supervision}: Without the visual auxiliary decoder forcing the latent tokens to encode spatial-temporal scene content, the latent representations collapse to encoding only language-level abstractions, losing the geometric precision needed for trajectory prediction. In the framing of the compression view of intelligence, these methods compress a symbolic abstraction of the world rather than the world itself---the compression target (language) is too abstract to drive the model toward genuinely causal scene representations. The visual decoder functions as a world model auxiliary: it provides a concrete, physically grounded compression target (future visual observations) that cannot be satisfied by language-level memorization. This is directly supported by our ablation. \OneVL{} w/o visual decoder scores 87.97, still below the full model (88.84), removing the world model supervision alone accounts for a $-$0.87 drop, even when all other components remain intact.
    
    \item \textbf{Absence of staged training}: Prior methods are typically optimized without \OneVL{}'s three-stage pipeline (\Cref{sec:training}), leaving language latents, visual latents, and trajectory prediction poorly aligned at the start of optimization. Our ablation isolates this effect. \OneVL{} w/o visual decoder but with staged training still reaches 87.97, well above the no-staged-training collapse (67.13) and above prior latent CoT baselines, demonstrating that the curriculum alone provides substantial gains independent of the visual decoder. Combining both yields the full 88.84, confirming that staged training and world model supervision are complementary. The curriculum creates a stable latent space that the world model decoder can then push toward genuinely causal representations.
\end{itemize}

To conclude, \OneVL{} addresses these failure modes as follows: (1) the visual auxiliary decoder provides precise spatial supervision and maintains spatial-temporal grounding in the latent representations; (2) the three-stage training recipe avoids the optimization conflict above by progressively aligning auxiliary decoders and latent tokens before full joint optimization.
\section{Conclusion}
\label{sec:conclusion}

We presented \OneVL{}, a framework for autonomous driving trajectory prediction built on a central hypothesis \emph{compression drives generalization}. A key contribution lies in identifying why prior latent CoT methods fail on planning tasks. Compressing language is not the same as compressing scene dynamics. Natural language descriptions of driving scenes are inherently abstract, encoding semantic labels rather than the physical causal structure that determines future outcomes. As a result, compressing language satisfies the efficiency requirement of the compression principle but not its intelligence requirement. \OneVL{} closes this gap by introducing a \emph{world model} auxiliary in the form of a visual decoder that predicts future-frame visual tokens. Future-frame prediction is a concrete, causally grounded compression target: a model that can anticipate the scene's visual evolution has necessarily internalized the dynamics governing agent motion and scene interaction, information that language alone cannot reliably encode.

At inference time, the decoders are discarded, and the prefilled latent tokens enable single-pass trajectory generation at a latency matching answer-only prediction. A three-stage training pipeline is essential for realizing this compression: Stage~0 establishes meaningful latent representations, Stage~1 aligns the decoders against a stable latent space, and Stage~2 tightens the bottleneck from both sides through bidirectional joint optimization. Ablations show that skipping this curriculum causes the compression to collapse into memorization, with a catastrophic performance drop.

Empirically, \OneVL{} achieves state-of-the-art results on NAVSIM, ROADWork, Impromptu, and Alpamayo-R1 while matching answer-only inference speed. Most notably, \OneVL{} is the only latent CoT method to outperform explicit autoregressive CoT, providing direct evidence that tighter compression, when grounded in both linguistic and world-model supervision, yields more effective reasoning than verbose token-by-token generation. Finally, by appending a lightweight MLP regression head, \OneVL{} retains competitive performance at only 5\% of the original latency, pointing toward efficient real-world deployment.

\paragraph{Limitations}
The current system requires roughly 3$\times$ memory during training, since three full 4B model instances must be held in memory. This is mitigated by DeepSpeed ZeRO-2 but still imposes nontrivial infrastructure requirements. In addition, the latent token count was chosen empirically. Thus, a systematic study of the trade-off between latent token count and representation capacity is left for future work.

\paragraph{Future Directions}
Several directions emerge naturally from this work. First, while \OneVL{}'s prefill mechanism eliminates latent CoT overhead, the trajectory tokens themselves are still generated autoregressively. 

As shown in ~\Cref{sec:eval:deployment}, appending an MLP head can improve inference speed at some cost to performance, so the overall latency remains constrained by AR decoding. 

Bridging this gap, for instance, through parallel or non-autoregressive trajectory decoding, is a key step toward true real-time deployment in safety-critical driving systems. Second, extending the world model decoder to multi-camera inputs would enable 360-degree future-scene prediction and more comprehensive causal scene understanding, further strengthening the compression targets available to the visual latents. Third, the dual-modal explanation framework could enable novel human-machine interface designs in which drivers receive real-time visual and verbal justifications for the vehicle's planning decisions. Beyond enhancing transparency and trust, this also has the potential to supply richer training signals for reinforcement learning, closing the loop between world model prediction and policy improvement.

\vspace{1cm}
\section{Contributions and Acknowledgments}
\label{sec:contributions}

\noindent
\begin{minipage}[t]{0.47\linewidth}
  \subsubsection*{Core Contributors}
  \begin{itemize}[leftmargin=*, itemsep=0.15em, topsep=0.25em]
    \item Jinghui Lu
    \item Jiayi Guan
    \item Zhijian Huang
    \item Jinlong Li
    \item Guang Li
    \item Lingdong Kong
    \item Yingyan Li
    \item Han Wang
    \item Shaoqing Xu
    \item Yuechen Luo
    \item Fang Li
    \item Chenxu Dang
    \item Junli Wang
    \item Tao Xu
    \item Jing Wu
    \item Jianhua Wu
    \item Xiaoshuai Hao
    \item Wen Zhang
    \item Tianyi Jiang
    \item Kuiyuan Yang
    \item Hangjun Ye
    \item Long Chen$^{\dagger}$
  \end{itemize}
\end{minipage}%
\hfill
\begin{minipage}[t]{0.47\linewidth}
  \subsubsection*{Contributors}
  \begin{itemize}[leftmargin=*, itemsep=0.15em, topsep=0.25em]
    \item Lingfeng Zhang
    \item Lei Zhou
    \item Yingbo Tang
    \item Jie Wang
    \item Yinfeng Gao
    \item Xizhou Bu
    \item Haochen Tian
    \item Yihang Qiu
    \item Feiyang Jia
    \item Lin Liu
    \item Yigu Ge
    \item Hanbing Li
    \item Yuannan Shen
    \item Jianwei Cui
    \item Hongwei Xie
    \item Bing Wang
    \item Haiyang Sun
    \item Jingwei Zhao
    \item Jiahui Huang
    \item Pei Liu
    \item Zeyu Zhu
    \item Yuncheng Jiang
    \item Zibin Guo
    \item Chuhong Gong
    \item Hanchao Leng
    \item Kun Ma
    \item Naiyan Wang
    \item Guang Chen
  \end{itemize}
\end{minipage}

\vspace{0.5em}
\noindent{\small\itshape $^{\dagger}$Corresponding author.}

\vspace{3em}
\noindent\textbf{Acknowledgments}

In this work, we utilize the Qwen3-VL-4B-Instruct~\cite{bai2025qwen3} model and the following datasets: NAVSIM~\cite{dauner2024navsim}, ROADWork~\cite{roadwork}, Impromptu~\cite{chiimpromptu}, and Alpamayo-R1~\cite{wang2025alpamayo}. The authors confirm that the use of these resources is strictly for academic research purposes and have not been involved in any commercial activities.

\vspace{0.4cm}
\bibliographystyle{plainnat}
\bibliography{main}

\clearpage\clearpage
\beginappendix

\section{Appendix}
\label{sec:appendix}

\subsection{Data Format Example}
\label{sec:appendix:data}

Each NAVSIM training sample in \OneVL{} follows the structure below. The assistant response contains visual latent tokens (4), language latent tokens (2), and the trajectory answer:

\begingroup
\setlength{\fboxsep}{8pt}%
\setlength{\fboxrule}{0.4pt}%
\noindent\color{black}\fbox{%
\begin{minipage}{\dimexpr\linewidth-2\fboxsep-2\fboxrule\relax}
\small
\sloppy
\setlength{\emergencystretch}{3em}
\color{black}\

\noindent{ User:}\ \detokenize{<image>}\ Front-view image of the driving scene.

\noindent{ Command:} MOVE FORWARD. Velocity: [1.5, 0.0]. Acceleration: [0.3, 0.0].

\noindent{ Historical trajectory:} (0.0, 0.0, 0.0), (-0.75, 0.0, 0.0), (-1.5, 0.0, 0.0).

\noindent\emph{[Task instruction to output reasoning and predicted trajectory\ldots]}

\medskip
\noindent{ Assistant:}

\noindent{
\detokenize{<|start-latent-vis|> <|latent-vis|> <|latent-vis|> <|latent-vis|> <|latent-vis|>}\\*
\detokenize{<|end-latent-vis|> <|start-latent|> <|latent|> <|latent|> <|end-latent|>}\\*
\detokenize{<answer>[0.75, 0.0, 0.0], [1.5, 0.0, 0.0], ...</answer>}}

\medskip
\noindent{ think\_steps} (label for $\mathcal{D}_\text{l}$): ``The road ahead is clear. There are no vehicles or pedestrians that would require braking. I should maintain current speed and continue straight along the lane.''

\medskip
\noindent{ future\_image\_tokens} (label for $\mathcal{D}_\text{v}$):\newline
{\detokenize{<|image_start|>...<|visual_token_XXXXXX|>...<|image_end|>}}
\end{minipage}%
}%
\endgroup

\subsection{Training Configuration}
\label{sec:appendix:training_config}

\noindent
We begin with self-supervised pretraining of the visual auxiliary decoder: the model learns to predict the next frame from ViT features alone, without the autoregressive backbone in the loop.
This phase uses \num{13040} optimizer steps and a global batch size of \num{256}, rather than a fixed epoch count.
We then move to \emph{Stage~0}, which warms up the main VLM while introducing latent tokens, followed by \emph{Stage~1}, where the auxiliary decoders train with the main model frozen and the visual decoder becomes action-conditioned through visual latent tokens.
Finally, \emph{Stage~2} performs joint end-to-end fine-tuning with all parts of the stack updated together; details are shown in Table~\ref{tab:training_config}.

\phantomsection
\subsection{CoT Annotation Construction}
\label{sec:appendix:roadwork_cot}
The chain-of-thought reasoning annotations for the ROADWork dataset are constructed using an in-house annotation pipeline developed specifically for work-zone driving scenarios. Unlike the NAVSIM annotations (sourced from AdaThinkDrive~\cite{adathinkdrive}), ROADWork CoT reasoning requires domain-specific annotation of:
\begin{itemize}
  \item Work-zone hazard identification: Detection and description of cones, barriers, temporary signage, and worker presence.
  \item Non-standard lane interpretation: Reasoning about temporary lane markings, reduced lane widths, and merge configurations.
  \item Speed and clearance reasoning: Justification of appropriate speed reduction and lateral clearance decisions specific to work zones.
\end{itemize}

The CoT reasoning annotations for the Impromptu VLA Dataset are constructed using a VLM-centric annotation pipeline with Chain-of-Thought prompting developed specifically for unstructured driving scenarios (the core corner cases for autonomous driving). Unlike annotations for conventional structured driving datasets, Impromptu VLA CoT reasoning requires domain-specific annotation of:

\begin{itemize}
    \item Unstructured scenario classification: Identification and categorization of four core challenging unstructured scenarios, including roads with unclear boundaries, temporary traffic rule changes, unconventional dynamic obstacles, and challenging road conditions.
    
    \item Complex scene element perception: Detection and description of ambiguous road boundaries, temporary traffic facilities, non-standard dynamic obstacles, and adverse road/environmental conditions.
    
    \item Non-standard driving behavior reasoning: Interpretation of temporary traffic rules, obstacle avoidance strategies, and justification of speed adjustment and lateral clearance decisions in unstructured environments.
    
    \item Planning-oriented decision reasoning: Generation of ego-vehicle meta-action plans, end-to-end trajectory predictions, and textual rationales for driving maneuvers.
\end{itemize}

Since the CoT reasoning labels for Alpamayo-R1 have not been released, we replicate these annotations ourselves. Specifically, we use the publicly released model checkpoint\footnote{\url{https://huggingface.co/nvidia/Alpamayo-R1-10B}} to reproduce the CoT labels for all training examples. Regarding the waypoint prediction, we applied a heuristic subsampling strategy to reduce the sequence from 64 to 8 points while ensuring the last points are retained, as 64 waypoints are too dense for autoregressive modeling (the original Alpamayo-R1 model employs flow matching). Furthermore, as the original paper does not release the official test set, we subsample 700 examples from the available video clips to construct our test set.

\noindent The following are four anonymized CoT examples:

\begingroup
\setlength{\fboxsep}{7pt}%
\setlength{\fboxrule}{0.4pt}%

\paragraph{NAVSIM Example}

\noindent\fbox{%
\begin{minipage}{\dimexpr\linewidth-2\fboxsep-2\fboxrule\relax}
\small
\noindent
The right side of the lane where the ego vehicle is located is close to the undrivable area, so I need to drive slightly to the left. I should pay more attention to a vehicle, located 18.42 meters ahead of the ego vehicle and 0.49 meters to the left. It is driving forward in the same direction, and its motion state is moving fast. Based on the understanding of the driving scene and the navigation information, the ego should maintain speed and keep the lane.
\end{minipage}%
}%

\medskip
\paragraph{ROADWork Example}

\noindent\fbox{%
\begin{minipage}{\dimexpr\linewidth-2\fboxsep-2\fboxrule\relax}
\small
\noindent
critical\_factors: An orange work-zone warning sign placed in the roadway ahead indicates construction activity and prompts caution and adjustment in path.
A line of cones along the left side of the roadway creates a temporary channelization/taper that narrows the usable lane and guides vehicles away from the left edge.
Parked vehicles tightly line the right curb, reducing lateral clearance and constraining the ego vehicle available path through the work area.
Scaffolding/sidewalk construction along the right side suggests active work frontage and encourages keeping distance from the curbside area.

cause:Taper or channelization speed adjustment and channelized lane keeping because cones taper the left side and parked vehicles along the right curb narrow the usable lane.

decision:longitudinal:Taper or channelization speed adjustment
lateral:Channelized lane keeping
\end{minipage}%
}%

\paragraph{Impromptu Example}

\noindent\fbox{%
\begin{minipage}{\dimexpr\linewidth-2\fboxsep-2\fboxrule\relax}
\small
\noindent
<decision>longitudinal decision: stop, lateral decision: straight</decision><cause> The decision to stop the car is likely due to the presence of construction barriers and a pedestrian crossing the road, indicating a need to ensure safety and comply with traffic regulations.</cause>
\end{minipage}%
}%

\paragraph{Alpamayo-R1 Example}

\noindent\fbox{%
\begin{minipage}{\dimexpr\linewidth-2\fboxsep-2\fboxrule\relax}
\small
\noindent
Turn right at the intersection since the right-turn traffic light is green.
\end{minipage}%
}%

\endgroup

\clearpage\clearpage
\subsection{LLM-as-Judge Evaluation Prompt}
\label{sec:appendix:llm_judge_prompt}

We employ a state-of-the-art proprietary VLM, gemini-3.1-flash-lite-preview\footnote{\url{https://ai.google.dev/gemini-api/docs/models/gemini-3.1-flash-live-preview}} as an automated evaluator for text CoT quality.
For each sample, the judge receives the front-camera image alongside the following prompt (with the ground-truth and predicted CoT filled in):

\begingroup
\setlength{\fboxsep}{7pt}%
\setlength{\fboxrule}{0.4pt}%
\noindent\fbox{%
\begin{minipage}{\dimexpr\linewidth-2\fboxsep-2\fboxrule\relax}
\small\sloppy
\setlength{\emergencystretch}{3em}
\rmfamily

You are an expert evaluator for Autonomous Driving Systems.\\
Your task is to evaluate the quality of a predicted driving Chain-of-Thought (CoT) against the Ground Truth CoT. You are also provided with the front-camera image for driving context.\\[4pt]
Ground Truth CoT: ``\{gt\_cot\}''\\
Predicted CoT: ``\{pred\_cot\}''\\[4pt]
Evaluation Criteria:\\
1.\ Perception (Distance \& Location): Minor numerical differences (e.g., 29.3m vs 27.5m) are acceptable and should only incur slight penalties.\\
2.\ Motion State (Prediction): Identifying the correct state (e.g., `moving fastly', `moving slowly', `keep static') is highly critical. A mismatch here is a severe safety hazard and must heavily reduce the score.\\
3.\ Ego Decision (Planning): The final action (e.g., `accelerate', `decelerate', `maintain speed', `keep lane') MUST match the Ground Truth. Any deviation is a critical safety error and should result in a massive penalty.\\
4.\ Language Fluency: Penalize minor grammatical errors.\\[4pt]
Score the Predicted CoT on a scale of 0 to 100 based on the above criteria.\\[4pt]
Output STRICTLY in the following JSON format without any markdown blocks, backticks, or extra text:\\
{\detokenize{{"reasoning": "Briefly explain the penalties applied based on the criteria.", "score": <integer_between_0_and_100>}}}
\end{minipage}%
}%
\endgroup

\noindent The judge's raw score (0--100) is normalized to $[0,\,1]$ by dividing by 100. We set the temperature to 0.1 to ensure scoring stability.

\begin{table*}[t]
    \centering
    \caption{\textbf{Training hyperparameters for \OneVL{}.}}
    \vspace{-0.1cm}
    \setlength{\tabcolsep}{5pt}
    \renewcommand{\arraystretch}{1.12}
    {\small %
    \begin{tabular}{@{}lcccc@{}}
        \toprule
        \textbf{Hyperparameter} & \textbf{Pre-training} & \textbf{Stage 0} & \textbf{Stage 1} & \textbf{Stage 2} \\
        \midrule
        \textbf{Steps} & \num{13040} & -- & -- & -- \\
        \textbf{Epochs} & -- & 2 & 1 & 5 \\
        \textbf{Batch (global)} & \num{256} & 64 & 64 & 64 \\
        \textbf{Learning rate} & $1{\times}10^{-4}$& $4{\times}10^{-5}$ & $1{\times}10^{-4}$ & $1{\times}10^{-4}$ \\
        \textbf{LR schedule} & Cosine & Cosine & Cosine & Cosine \\
        \textbf{Optimizer} & AdamW & AdamW & AdamW & AdamW \\
        \textbf{Precision} & BF16 & BF16 & BF16 & BF16 \\
        \textbf{Parallelism} & Zero-2 & Zero-2 & Zero-2 & Zero-2 \\
        \midrule
        \textbf{Trainable} & Vis aux decoder & ViT, LLM, aligner & Lang.\ \& vis aux dec & All \\
        \textbf{Frozen} & --  & -- & Main VLM & -- \\
        \midrule
        \rowcolor[HTML]{FFE0CC}
        \textbf{$\lambda_{l}$} & -- & -- & 1.0 & 1.0 \\
        \rowcolor[HTML]{FFE0CC}
        \textbf{$\lambda_{v}$} & 1.0 & -- & 0.1 & 0.1 \\
        \bottomrule
    \end{tabular}%
    }
    \label{tab:training_config}
    \vspace{0.1cm}
\end{table*}

\begin{table*}[t]
\centering
\caption{\textbf{Performance comparisons on the Impromptu benchmark} \cite{chiimpromptu}. We report trajectory prediction L2 error following the benchmark setting (meters; lower is better).}
\vspace{-0.1cm}
\setlength{\tabcolsep}{15pt}
\begin{tabular}{lcccccc}
\toprule
\multicolumn{6}{c}{\textbf{Traj. Pred. L2 Error (m)}} \\
\textbf{Method} & \textbf{1s} $\downarrow$ & \textbf{2s} $\downarrow$ & \textbf{3s} $\downarrow$ & \textbf{4s} $\downarrow$  & \textbf{Avg.} $\downarrow$ \\
\midrule
$\circ$~Impromptu VLA~\cite{chiimpromptu}* &0.90 & 2.80  & 3.75 & 5.89 & 3.16 \\
$\circ$~Impromptu VLA~\cite{chiimpromptu} &0.14 & 0.60  & 1.45 & 2.67 & 1.22 \\
\rowcolor[HTML]{FFE0CC}
$\bullet$~\textbf{OneVL}           &  \textbf{0.13} & \textbf{0.48} & \textbf{1.18} & \textbf{2.25} & \textbf{1.01} \\
\bottomrule
\end{tabular}
\label{tab:impromptu_reproduce_appendix}
\vspace{0.2cm}
\end{table*}

\FloatBarrier
%
\subsection{Reproducing Impromptu}

To reproduce the optimal performance of the baseline model, we follow the official settings to conduct evaluations using both the training script released by Impromptu and the pre-trained checkpoint uploaded to Hugging Face, with the corresponding test results presented in Table~\ref{tab:impromptu_reproduce_appendix}. Concretely, the first row marked with an asterisk (*) reports the results evaluated on the author-provided checkpoint\footnote{\url{https://huggingface.co/aaaaaap/ImpromptuVLAModel/tree/main/7B_AD}}, and the second row shows the outcomes of the model trained from scratch via the official training script. As neither of these results matches the performance claimed in the original paper, we adopt the better-performing one as the representative result of Impromptu for comparative analysis in the main experiments.

\subsection{NAVSIM qualitative examples}
\label{sec:appendix:navsim_qualitative}
\begingroup
\captionsetup{font=footnotesize,skip=3pt}

\noindent Figure~\ref{fig:navisim_example2} to \ref{fig:navisim_example10} show additional NAVSIM examples compare a baseline answer-only prediction against \OneVL{}, along with two future frames decoded from the visual auxiliary decoder, the text CoT from the language decoder. The key reasoning that affects the decision is highlighted in bold. Each plot overlays ground-truth (\textcolor{green!70!black}{green}) and predicted (\textcolor{red!70!black}{red}) trajectories on the front camera view.
\FloatBarrier

\endgroup

\FloatBarrier
%
\providecommand{\RoadworkQualFigRoot}{figs/roadwork_vis_e4}
\providecommand{\RoadworkQualColW}{0.28\linewidth}
\providecommand{\RoadworkQualLblW}{0.28\linewidth}
\providecommand{\RoadworkQualTabGap}{0.014\linewidth}
\providecommand{\RoadworkQualA}{\RoadworkQualFigRoot/example_29_idx1419}
\providecommand{\RoadworkQualB}{\RoadworkQualFigRoot/example_14_idx1457}

\subsection{Roadwork qualitative examples}
\label{sec:appendix:roadwork_qualitative}
\begingroup
\captionsetup{font=footnotesize,skip=3pt}

\noindent Figure~\ref{fig:roadwork_example2} to~\ref{fig:roadwork_example3} show additional ROADWork examples compare a baseline answer-only prediction against \OneVL{}, along with two future frames decoded from the visual auxiliary decoder, the text CoT from the language decoder. Each plot overlays ground-truth (\textcolor{green!70!black}{green}) and predicted (\textcolor{red!70!black}{red}) trajectories on the front camera view.

\subsection{Impromptu qualitative examples}
\label{sec:appendix:impromptu_qualitative}
\begingroup
\captionsetup{font=footnotesize,skip=3pt}

Figures~\ref{fig:nuScences_678_8198} and~\ref{fig:nuScences_927_10990} present the cases of the answer-only baseline and the OneVL model on the impromptu task. Green lines in the figures denote the ground-truth trajectories provided by the dataset, and red lines represent the trajectories predicted by different models. As observed in Figure~\ref{fig:nuScences_678_8198}, the CoT prompt guides the model to avoid large vehicles merging into the traffic flow, enabling OneVL to generate trajectories that stay away from large vehicles and are closer to the ground truth. Furthermore, Figure~\ref{fig:nuScences_927_10990} shows that when passing through a Y-shaped intersection, the right-turn CoT prompt allows OneVL to output a reasonable right-turn trajectory that remains more centered in the lane. Comprehensive analysis demonstrates that the information provided by CoT prompts can help produce trajectories that are closer to the ground truth and better satisfy driving requirements, especially in scenarios that require changes in current driving behavior and trajectories, such as crossing intersections and interacting with surrounding vehicles.

\begin{figure}[t]
    \centering
    \includegraphics[width=\textwidth]{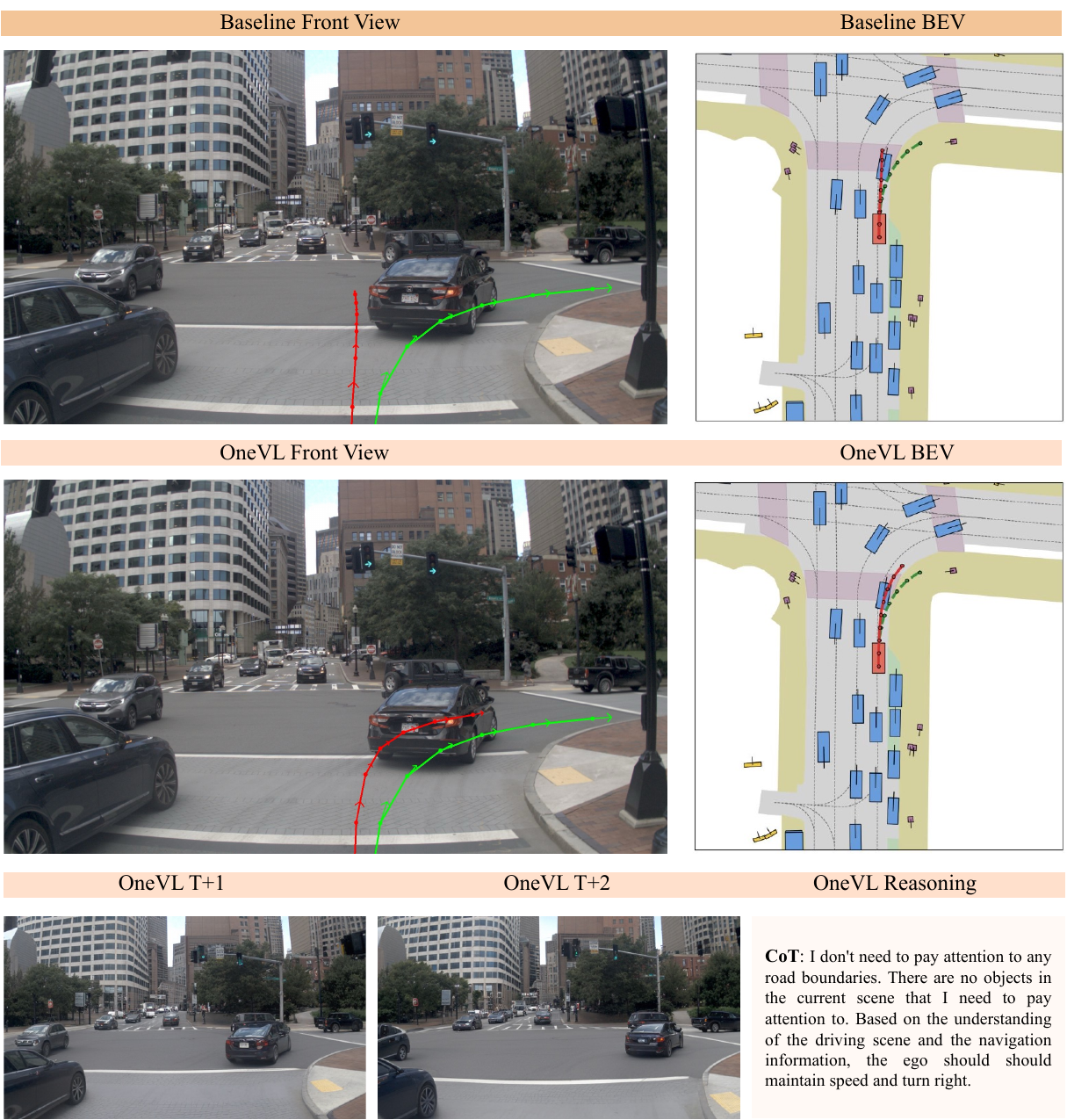}
    \vspace{0.1cm}
    \caption{\textbf{NAVSIM qualitative example 1.}}
\label{fig:navisim_example2}
\end{figure}

\begin{figure}[t]
    \centering
    \includegraphics[width=\textwidth]{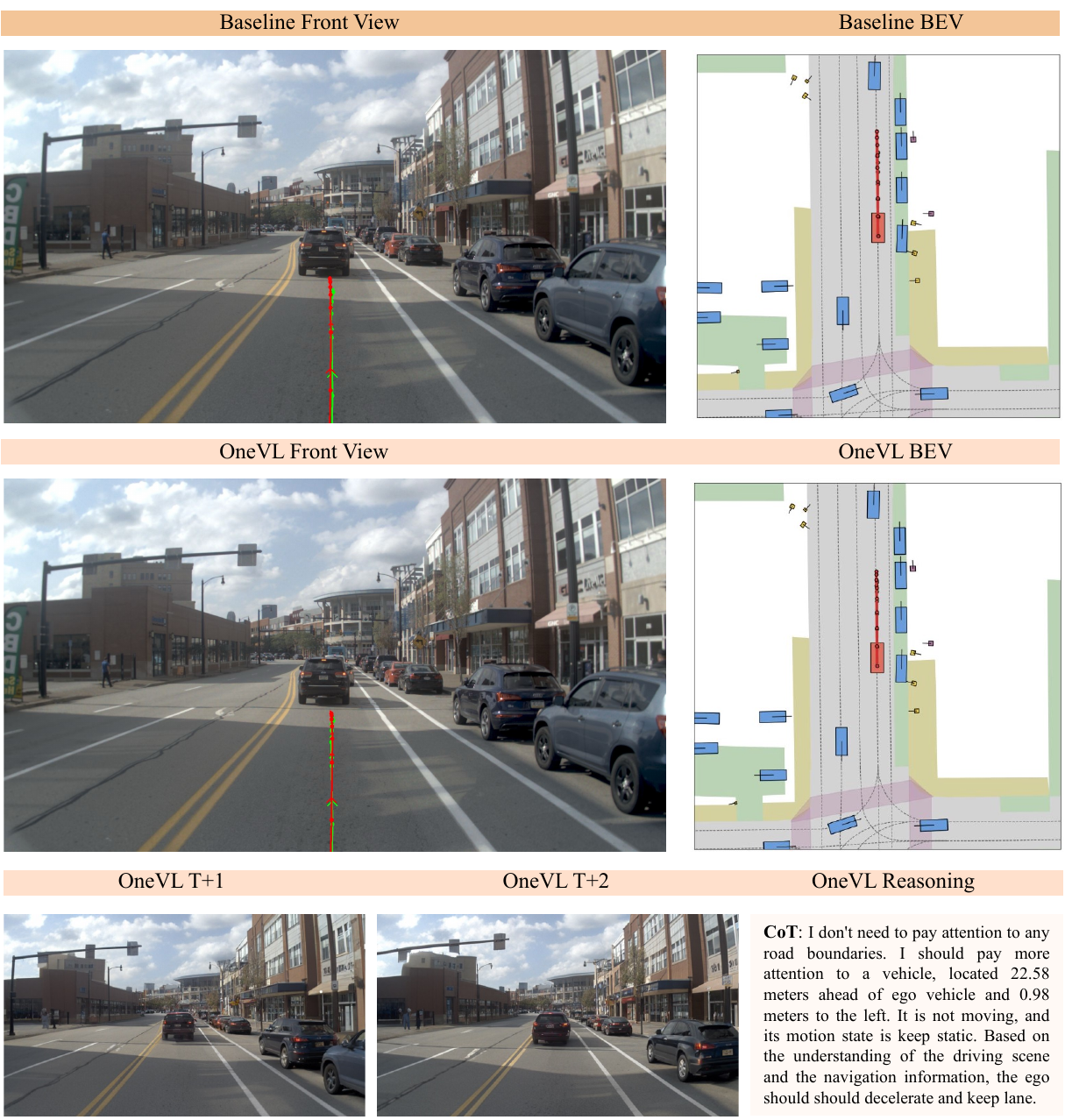}
    \vspace{0.1cm}
    \caption{\textbf{NAVSIM qualitative example 2.}}
\label{fig:navisim_example3}
\end{figure}

\begin{figure}[t]
    \centering
    \includegraphics[width=\textwidth]{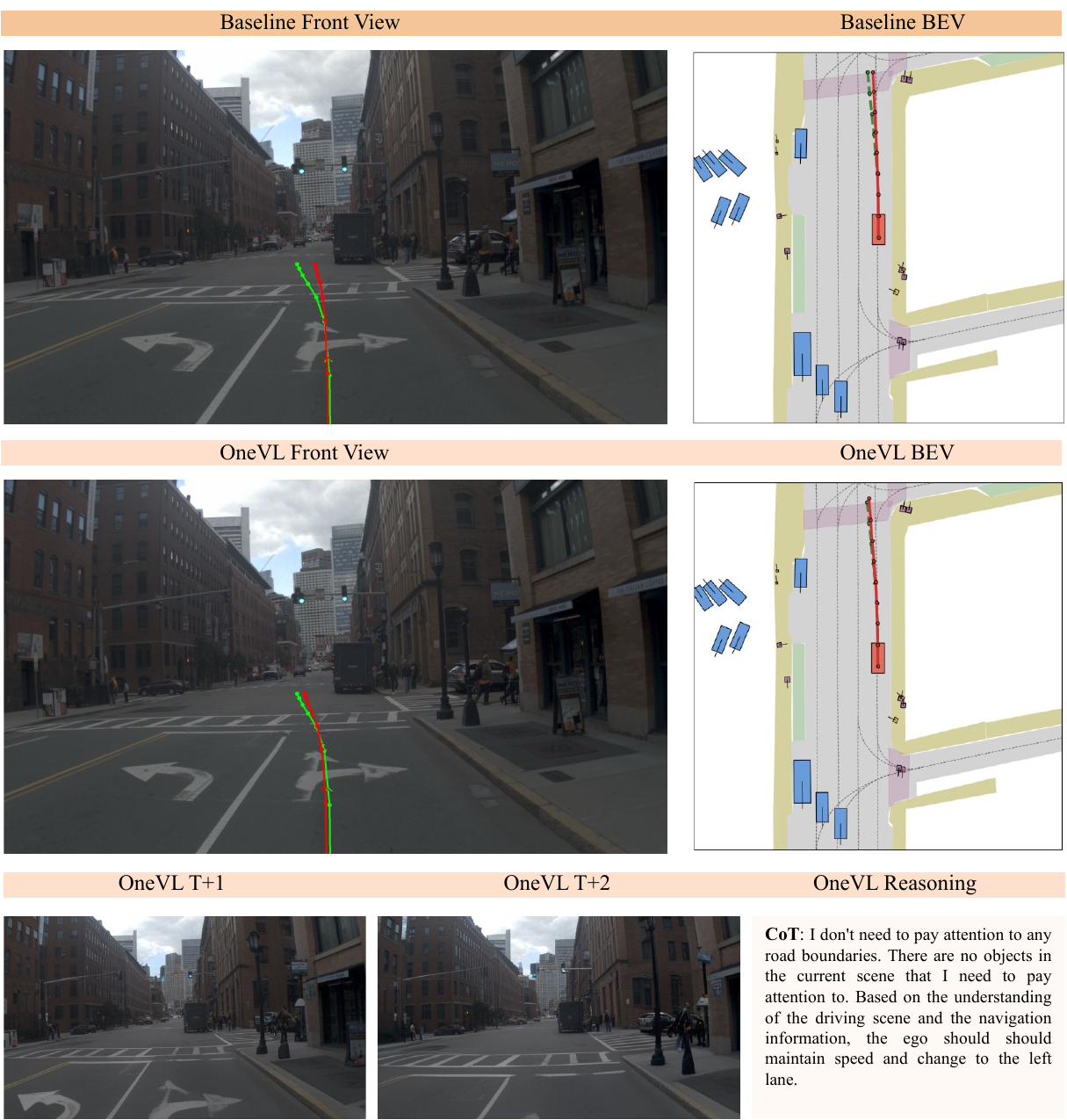} 
    \vspace{0.1cm}
    \caption{\textbf{NAVSIM qualitative example 3.}}
\label{fig:navisim_example4}
\end{figure}

\begin{figure}[t]
    \centering
    \includegraphics[width=\textwidth]{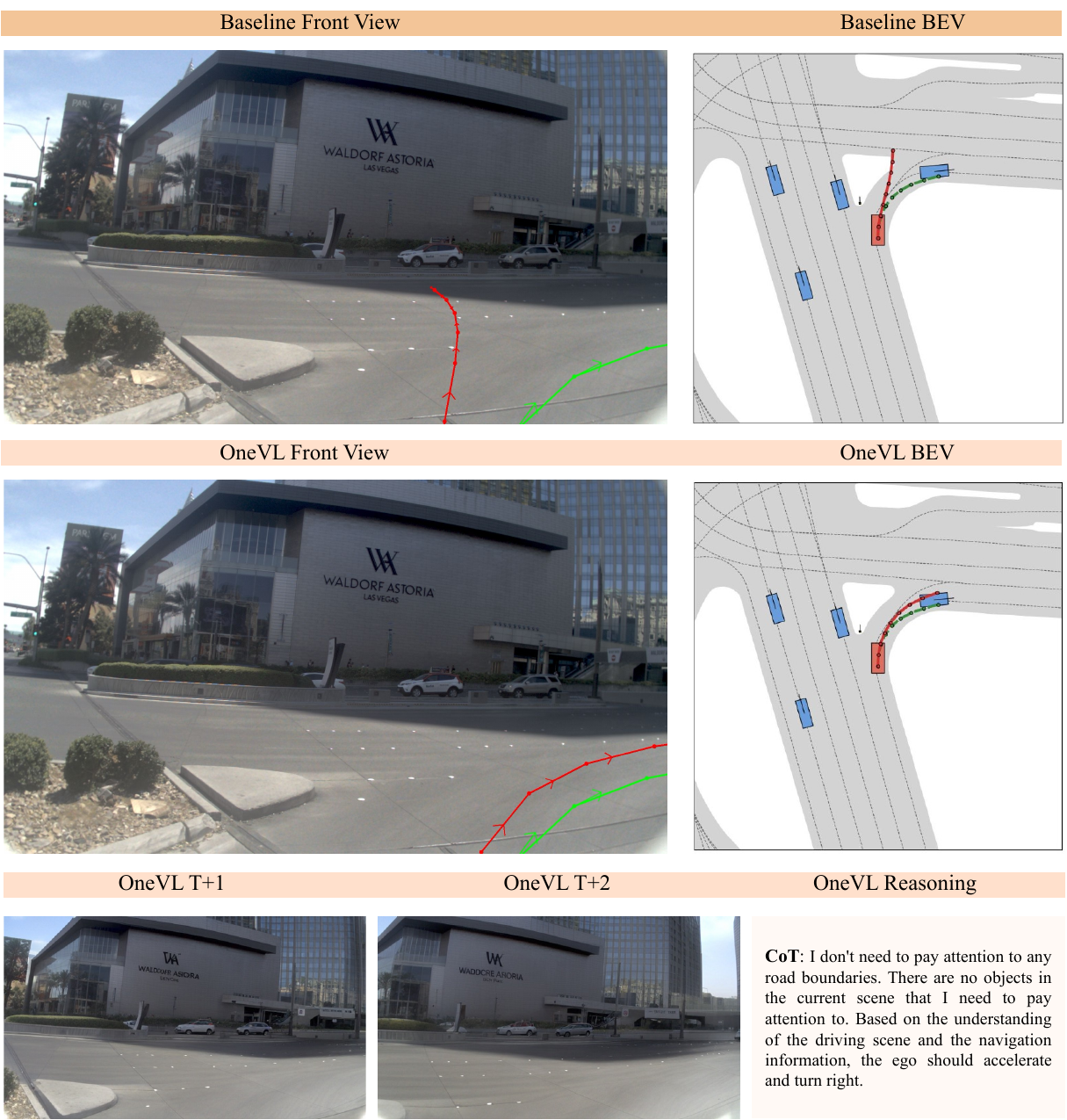} 
    \vspace{0.1cm}
    \caption{\textbf{NAVSIM qualitative example 4.}}
\label{fig:navisim_example5}
\end{figure}

\begin{figure}[t]
    \centering
    \includegraphics[width=\textwidth]{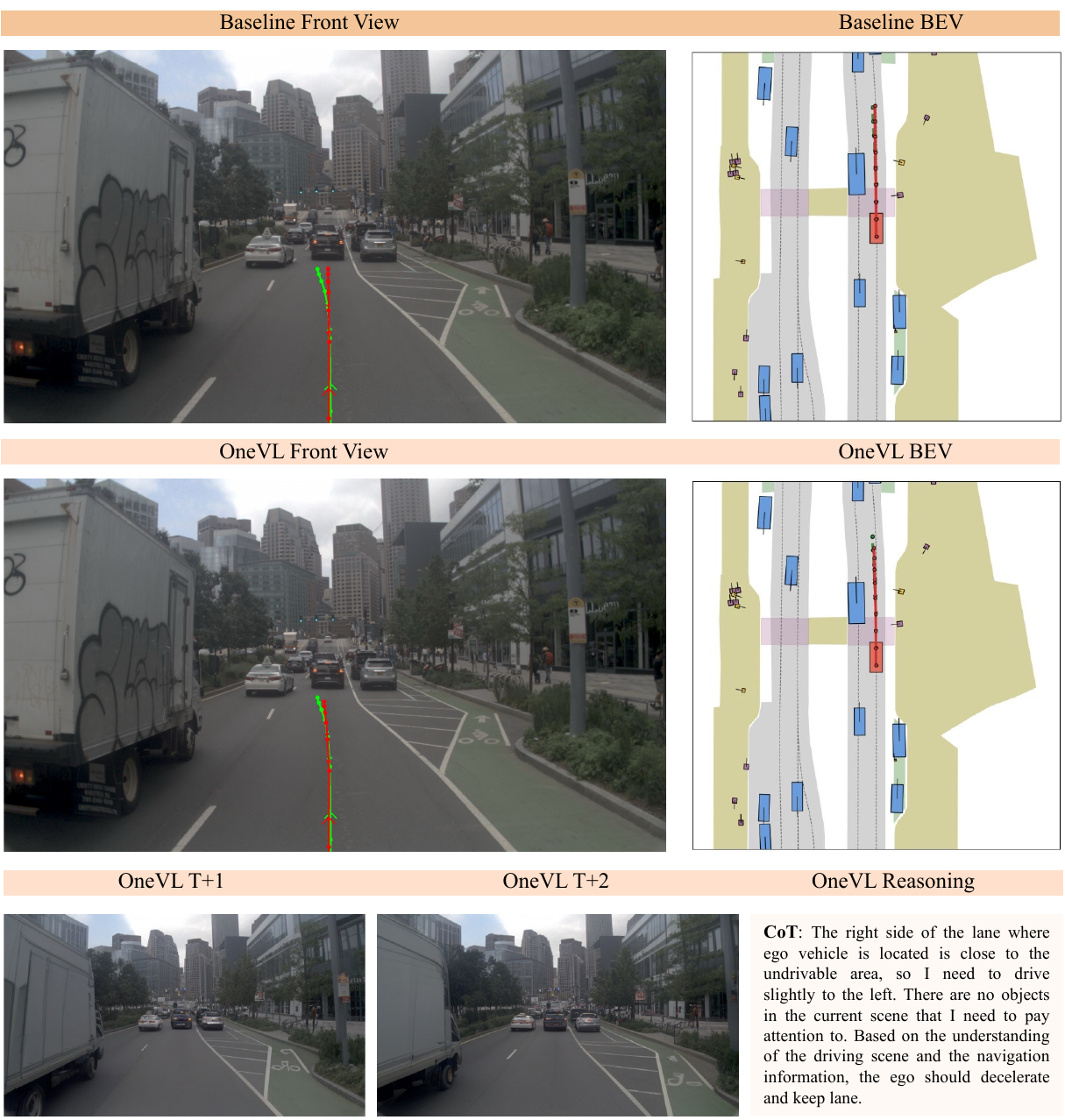} 
    \vspace{0.1cm}
    \caption{\textbf{NAVSIM qualitative example 5.}}
\label{fig:navisim_example6}
\end{figure}

\begin{figure}[t]
    \centering
    \includegraphics[width=\textwidth]{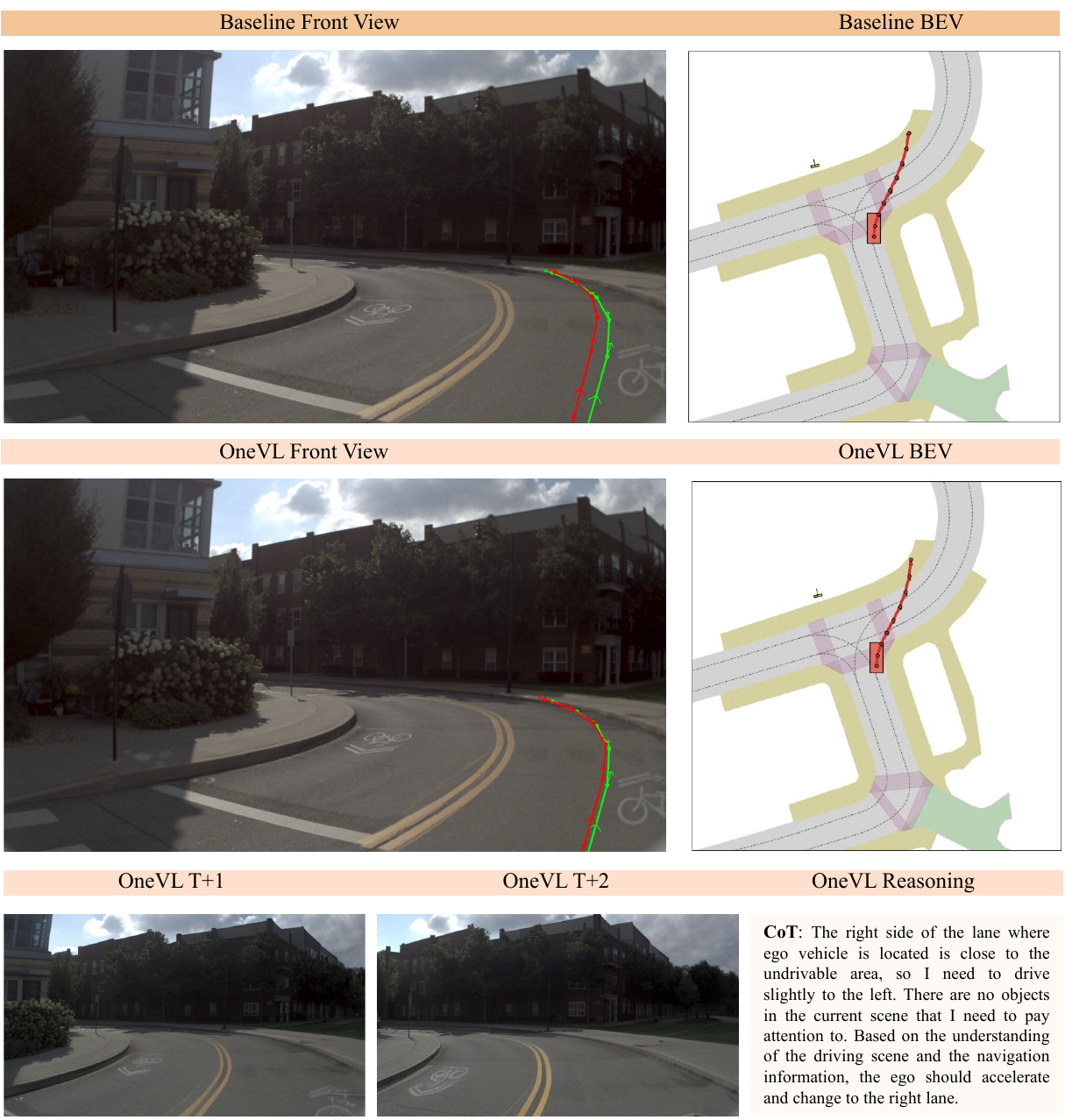} 
    \vspace{0.1cm}
    \caption{\textbf{NAVSIM qualitative example 6.}}
\label{fig:navisim_example7}
\end{figure}

\begin{figure}[t]
    \centering
    \includegraphics[width=\textwidth]{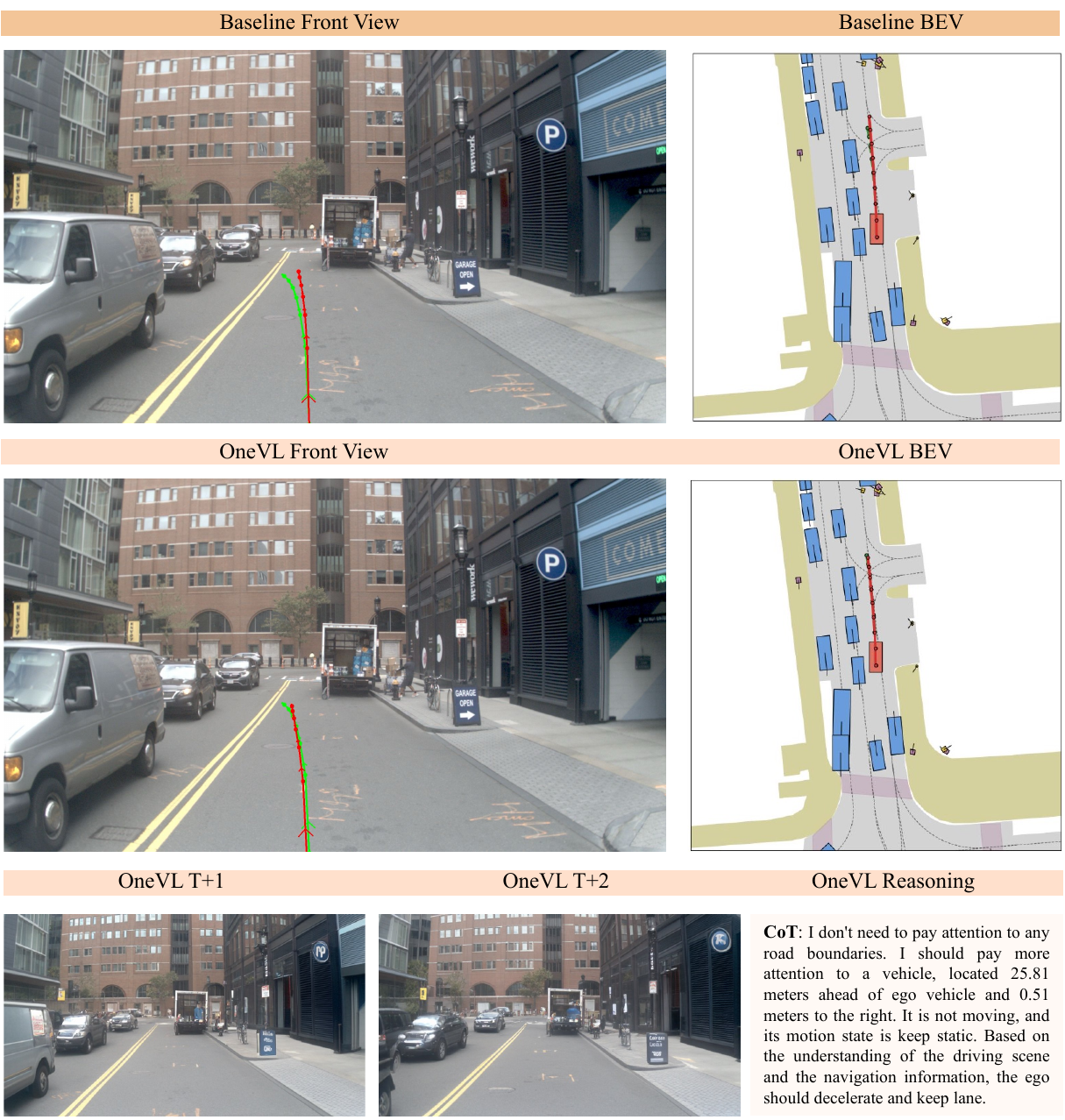} 
    \vspace{0.1cm}
    \caption{\textbf{NAVSIM qualitative example 7.}}
\label{fig:navisim_example8}
\end{figure}

\begin{figure}[t]
    \centering
    \includegraphics[width=\textwidth]{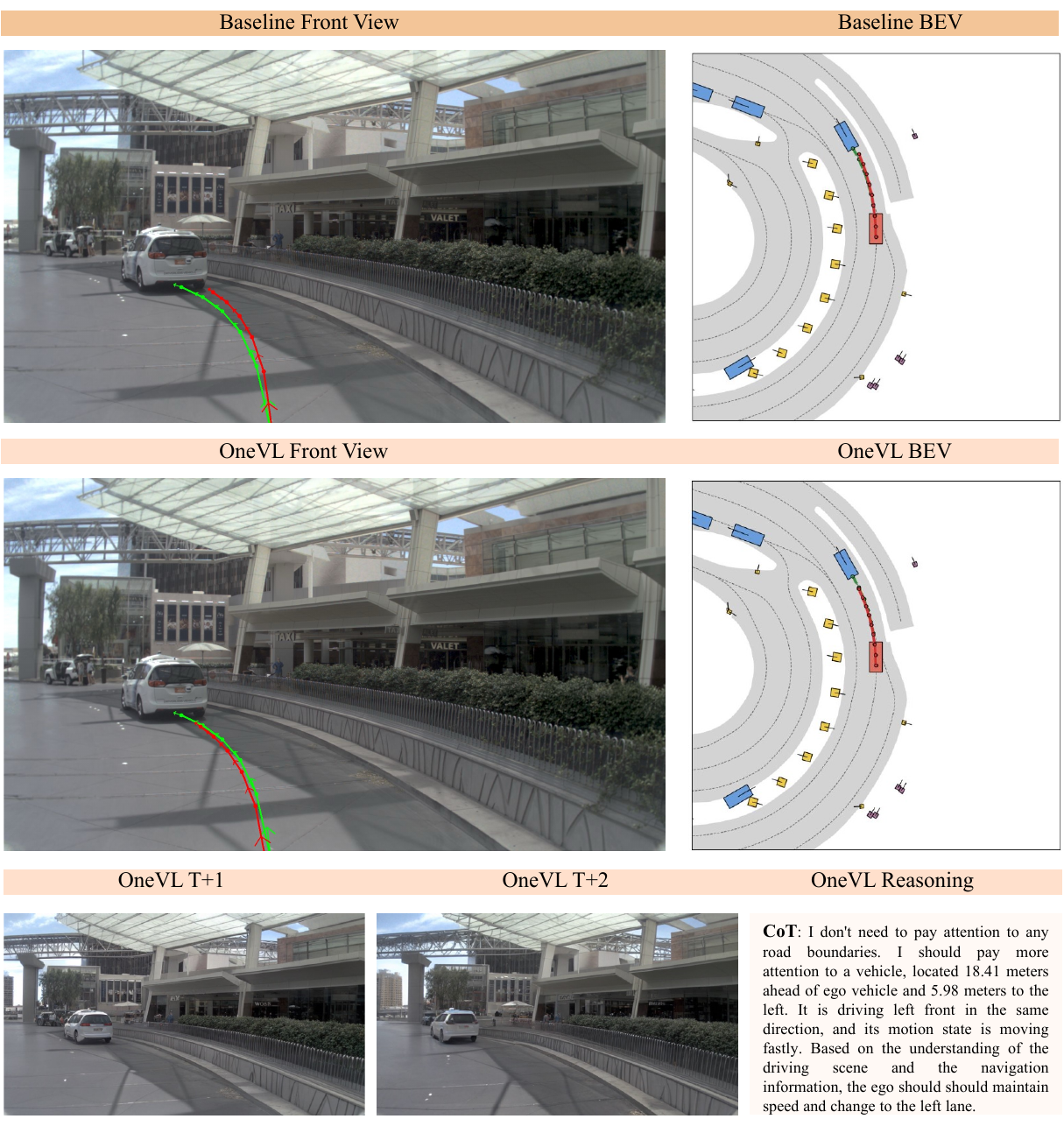}
    \vspace{0.1cm}
    \caption{\textbf{NAVSIM qualitative example 8.}}
\label{fig:navisim_example9}
\end{figure}

\begin{figure}[t]
    \centering
    \includegraphics[width=\textwidth]{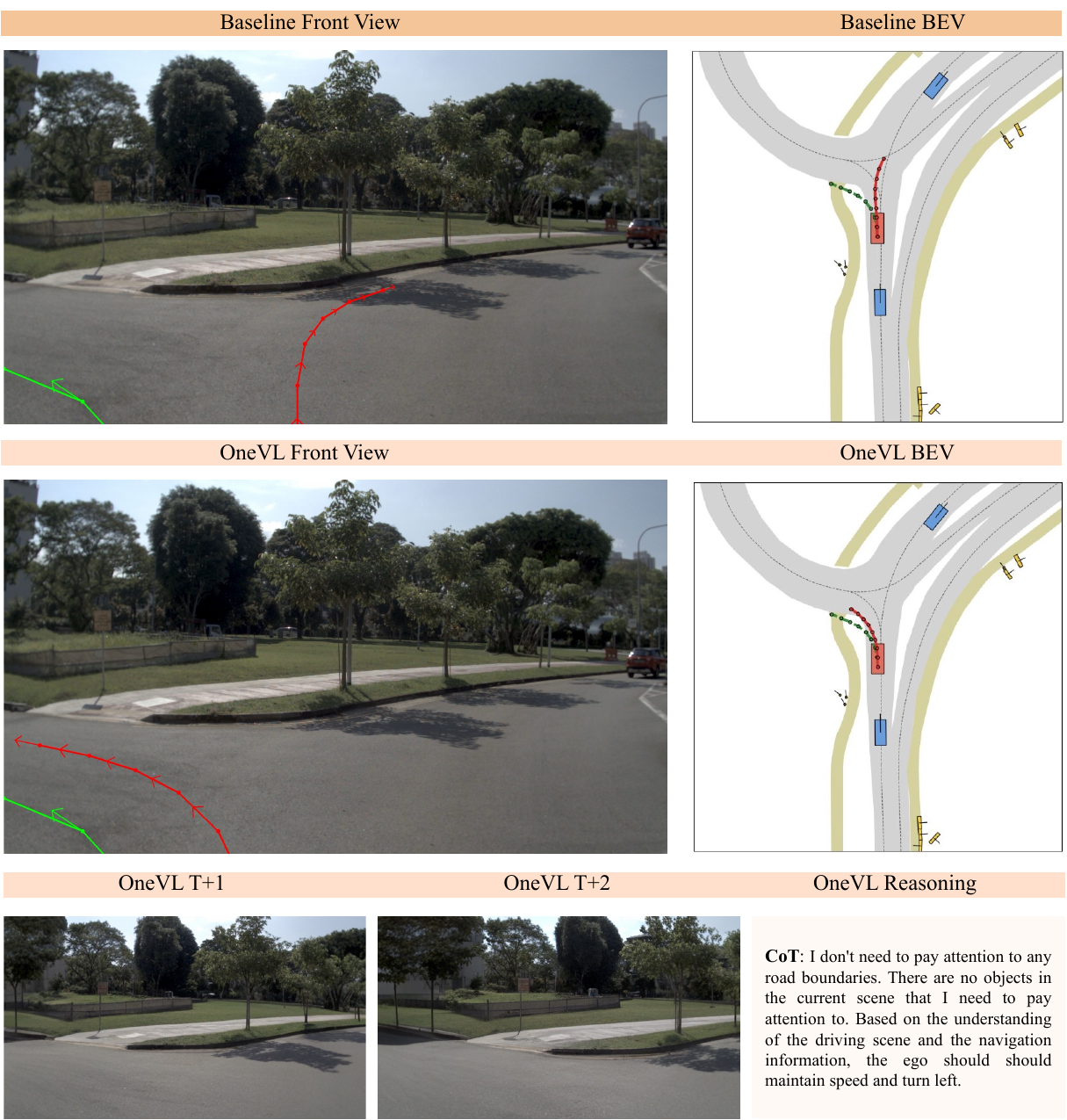} 
    \vspace{0.1cm}
    \caption{\textbf{NAVSIM qualitative example 9.}}
\label{fig:navisim_example10}
\end{figure}

\begin{figure}[t]
    \centering
    \includegraphics[width=\textwidth]{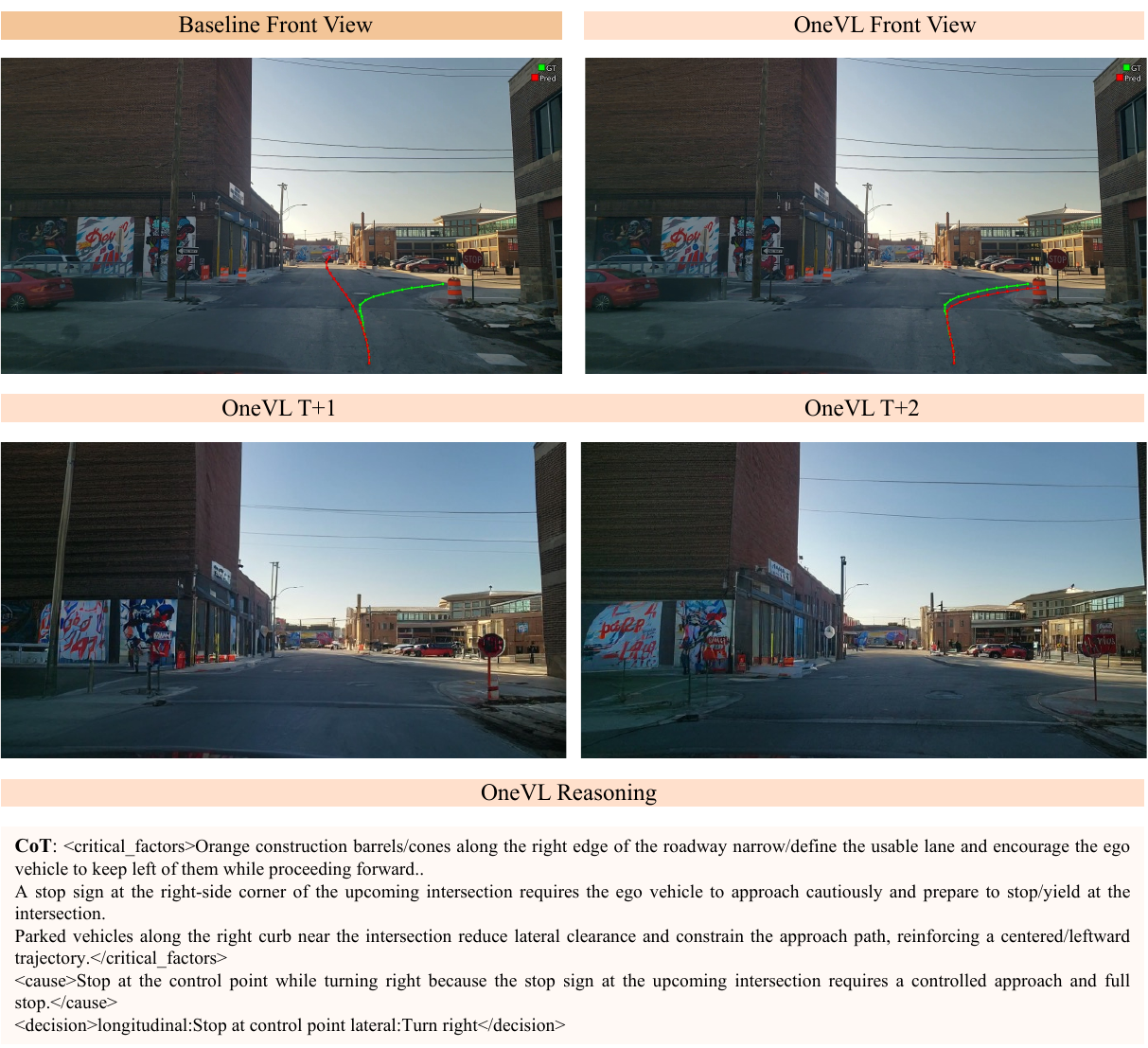} 
    \vspace{0.1cm}
    \caption{\textbf{ROADWork qualitative example 1.}}
\label{fig:roadwork_example2}
\end{figure}

\begin{figure}[t]
    \centering
    \includegraphics[width=\textwidth]{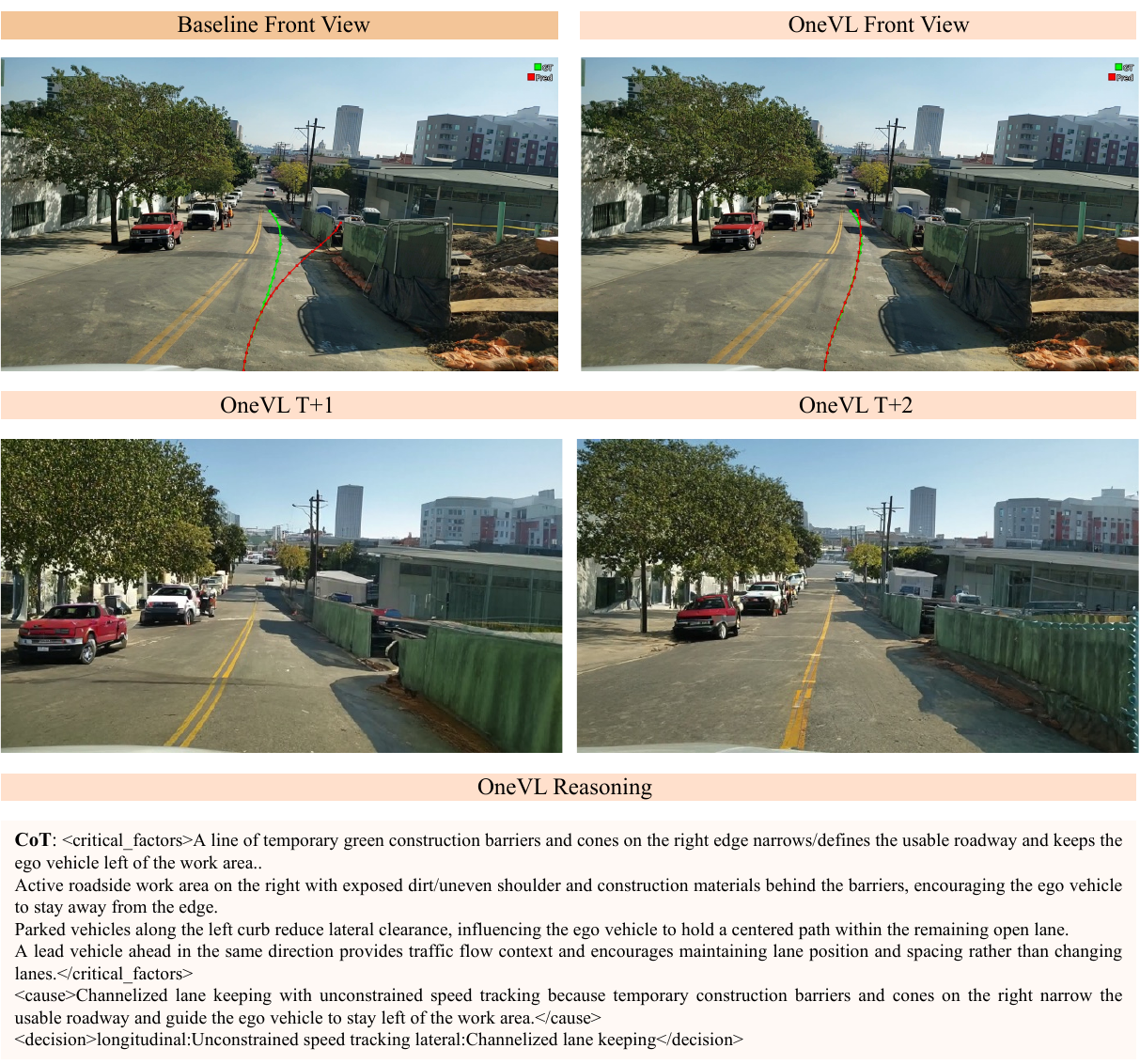}
    \vspace{0.1cm}
    \caption{\textbf{ROADWork qualitative example 2.}}
\label{fig:roadwork_example3}
\end{figure}

\begin{figure}[t]
    \centering
    \includegraphics[width=\textwidth]{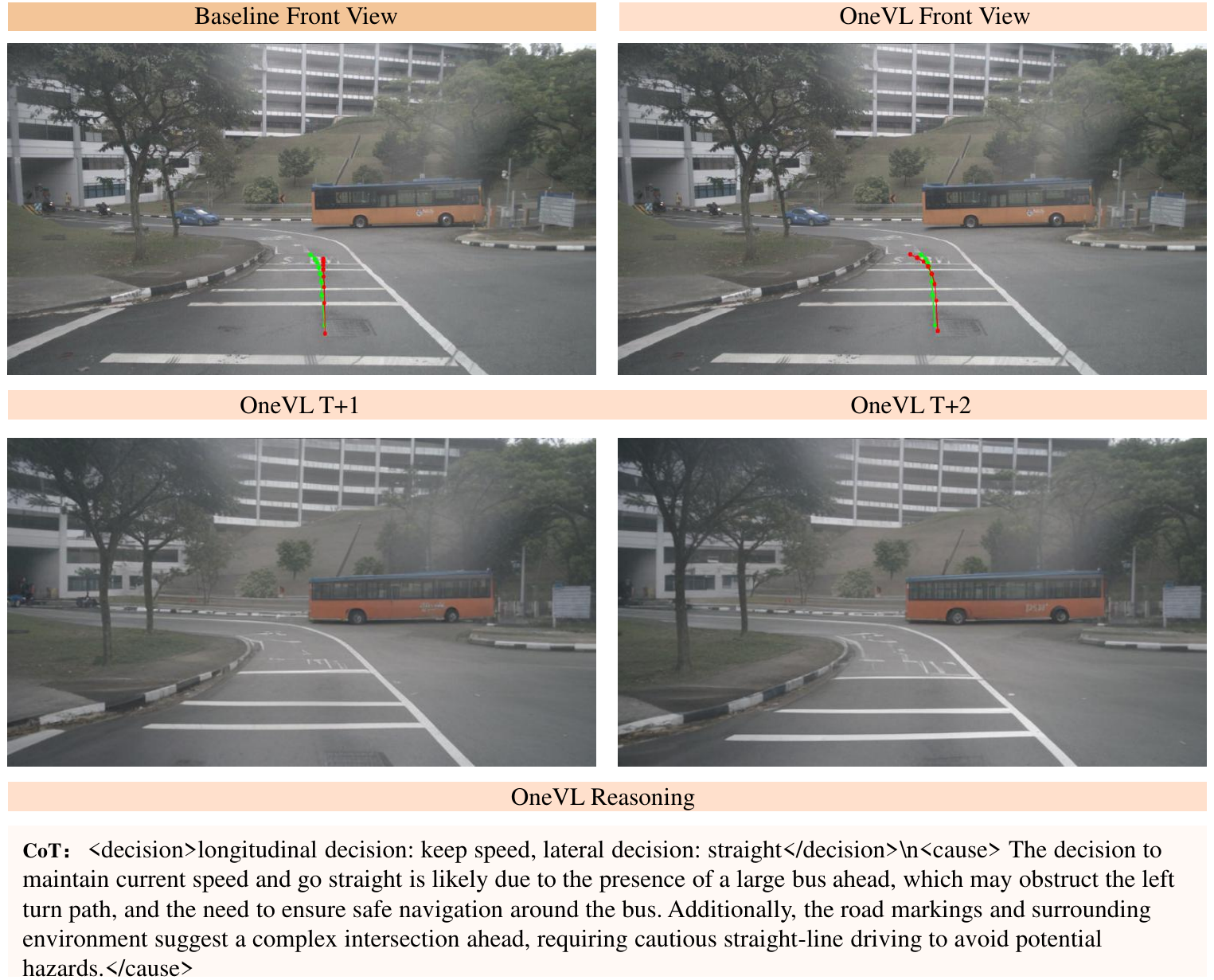} 
    \vspace{0.1cm}
    \caption{\textbf{Impromptu qualitative example 1.}}
\label{fig:nuScences_678_8198}
\end{figure}

\begin{figure}[t]
    \centering
    \includegraphics[width=\textwidth]{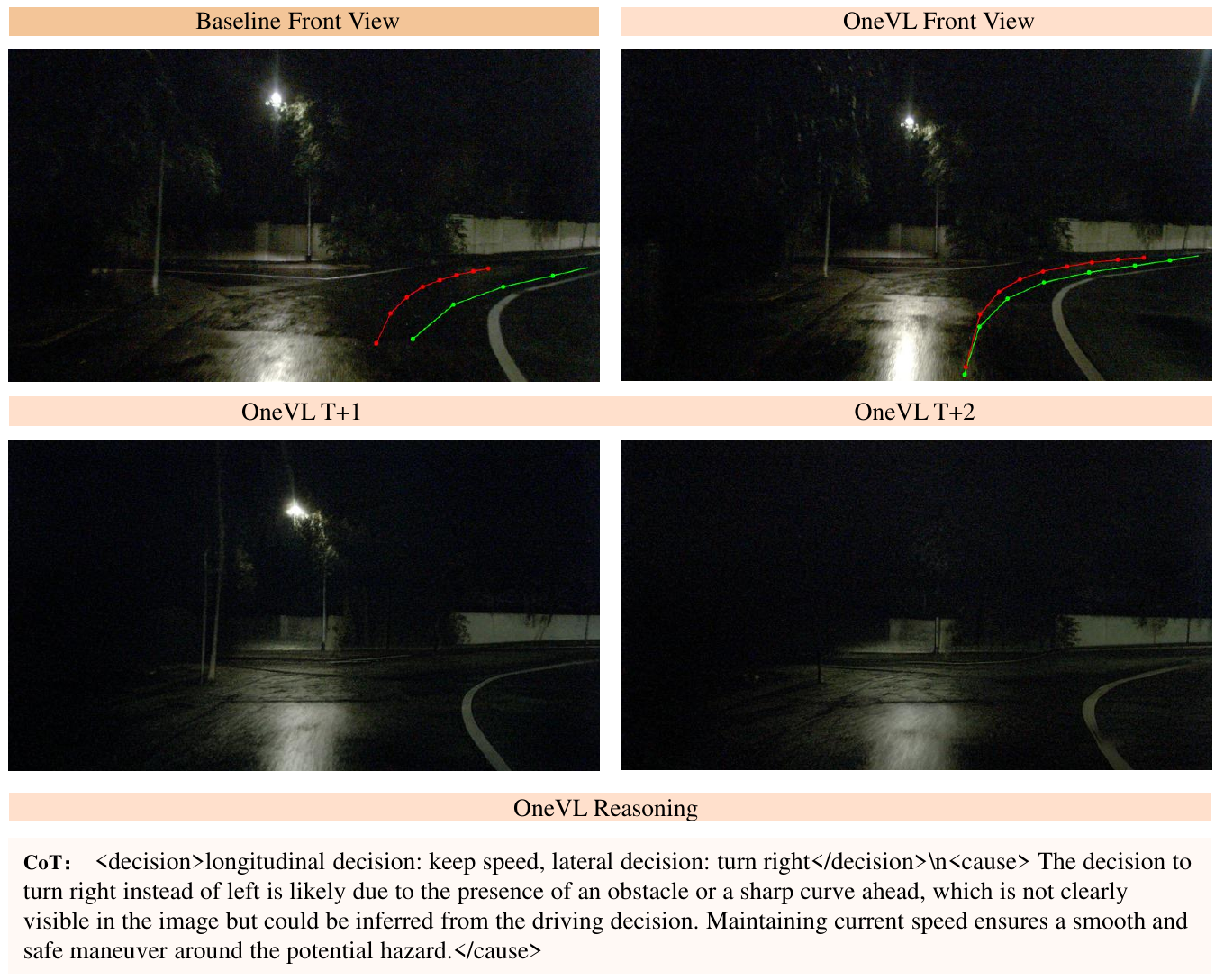}
    \vspace{0.1cm}
    \caption{\textbf{Impromptu qualitative example 2.}}
\label{fig:nuScences_927_10990}
\end{figure}

\subsection{Alpamayo-R1 qualitative examples}
\label{sec:appendix:apr1_qualitative}
\begingroup
\captionsetup{font=footnotesize,skip=3pt}

Figure~\ref{fig:ar1_appendix_example1} presents the cases of the answer-only baseline and the OneVL model on the Alpamayo-R1 dataset.

\begin{figure}[t]
    \centering
    \includegraphics[width=\textwidth]{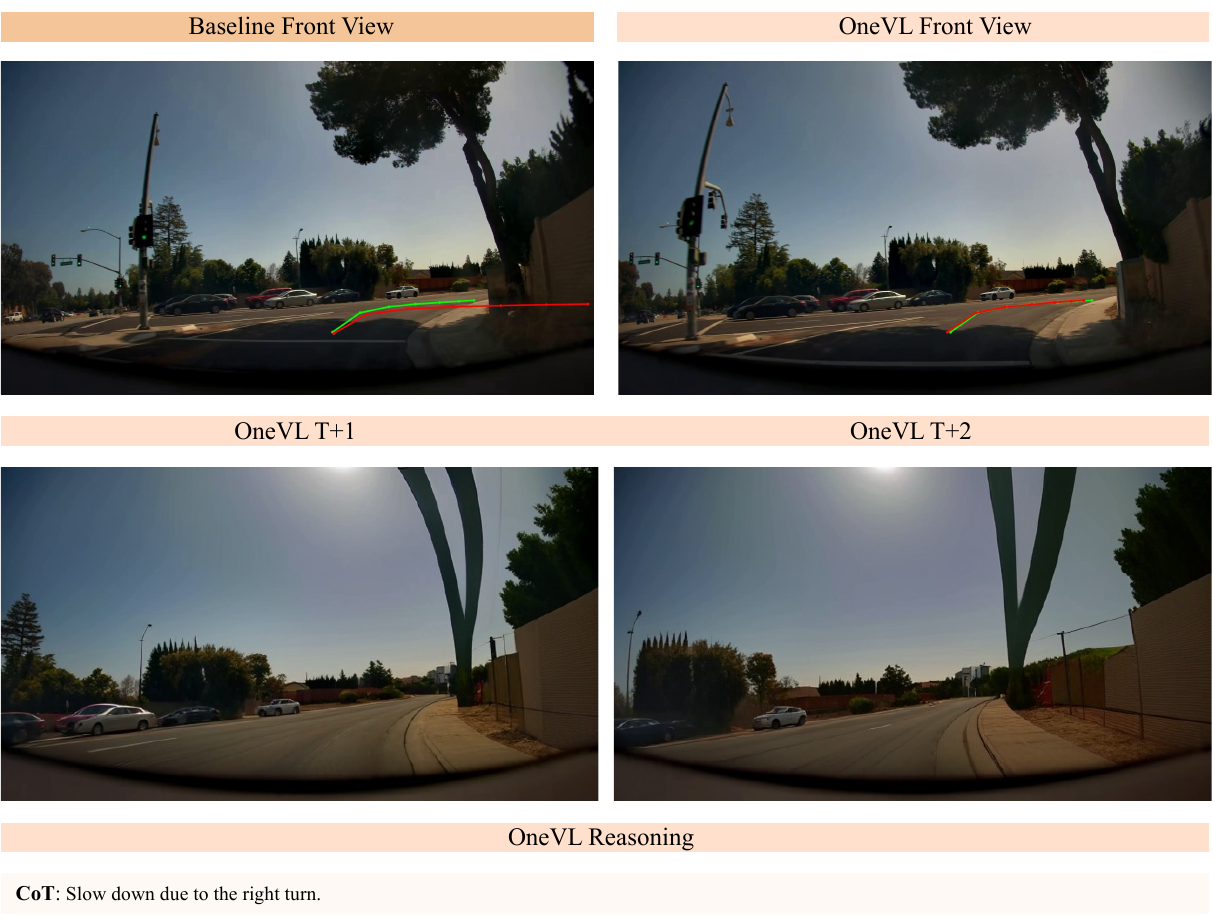} 
    \vspace{0.1cm}
    \caption{\textbf{Alpamayo-R1 qualitative example.}}
\label{fig:ar1_appendix_example1}
\end{figure}

\FloatBarrier
\endgroup
\clearpage
\let\clearpage\relax

\end{document}